\documentclass[runningheads]{llncs}

\usepackage{eccv}

\usepackage{eccvabbrv}

\usepackage{graphicx}
\usepackage{booktabs}
\usepackage{enumitem}
\usepackage{multirow}
\usepackage[linesnumbered,ruled]{algorithm2e}
\usepackage{wrapfig}
\usepackage{xcolor}

\usepackage[accsupp]{axessibility}  

\usepackage{hyperref}
\usepackage[capitalize]{cleveref}
\usepackage{orcidlink}

\begin{document}

\title{Multi-Channel Uncertainty-Weighted Score Matching for Conditional Diffusion in Medical UDA} 

\titlerunning{UPDiff-UDA}

\author{Chen Li\thanks{Email: Chen Li (li.chen.8@stonybrook.edu).}\inst{1} \and
Meilong Xu\inst{1}\ \and
Xiaoling Hu\inst{2} \and 
Weimin Lyu\inst{1} \and
Chao Chen\inst{1} }


\institute{Stony Brook University, Stony Brook, NY, USA \and
Massachusetts General Hospital and Harvard Medical School, MA, USA}

\maketitle

\begin{abstract}
Robust medical image segmentation across modalities remains challenging due to severe domain shifts and the lack of target-domain labels. While diffusion models have been explored for cross-domain generation and augmentation, target-domain conditional diffusion training typically relies on highly noisy pseudo masks; naively conditioning on a single Arg-Max pseudo-label can corrupt diffusion training and downstream segmentation. We propose \textbf{UPDiff-UDA}, a unified UDA framework whose core is an uncertainty-guided training objective for target-domain conditional diffusion. Given an imperfect source-trained segmenter, we use its per-pixel softmax distribution to form ranked pseudo-label maps (Arg-Max, Arg-2nd, Arg-3rd, \dots). Each map yields a conditional score estimate, and we aggregate them via pixel-wise confidence weighting to obtain an uncertainty-reweighted score for score matching, improving robustness to pseudo-label noise while leveraging alternative plausible labels in uncertain regions. We further provide a theoretical justification showing that confidence-weighted aggregation follows a minimum-MSE convex-combination principle under the segmenter-induced surrogate label distribution. To improve pseudo-condition quality, we also introduce a feature-guided, low-degree-of-freedom Bézier curve adaptation to reduce appearance gaps. Experiments on multiple public datasets and modality shifts show that UPDiff-UDA generates high-fidelity labeled target-style samples for augmentation and consistently outperforms strong UDA baselines. The code for this project is available at:
\url{https://github.com/superlc1995/Multi-Channel-Uncertainty-Diffusion-UDA}
\end{abstract}

\section{Introduction}
\label{sec:intro}

Deep learning models can achieve excellent segmentation performance when trained with dense annotations, but their reliability often collapses under domain shift, e.g., when imaging protocols, scanners, or institutions change. In medical imaging, such shifts are very common and can degrade well-trained models from another domain. Unsupervised domain adaptation (UDA)~\cite{chen2019synergistic, huo2018synseg, jiang2020psigan, liu2019susan} aims to transfer knowledge from a labeled source domain to an unlabeled target domain, reducing the need for expensive target annotations. GAN-based style translation is a popular UDA approach~\cite{liu2017unsupervised, zhu2017unpaired}, either mapping target images to source style for inference with a source-trained model or translating source images to target style for data augmentation. However, beyond GAN training instability, such translations often degrade or distort highly variable regions, especially small, rare lesions with heterogeneous appearance, because reliable cross-domain correspondences and consistent style mappings are difficult to learn~\cite{wang2024cyclesgan}.

\begin{figure*}[t]
    \centering 
   \includegraphics[width=1\linewidth]{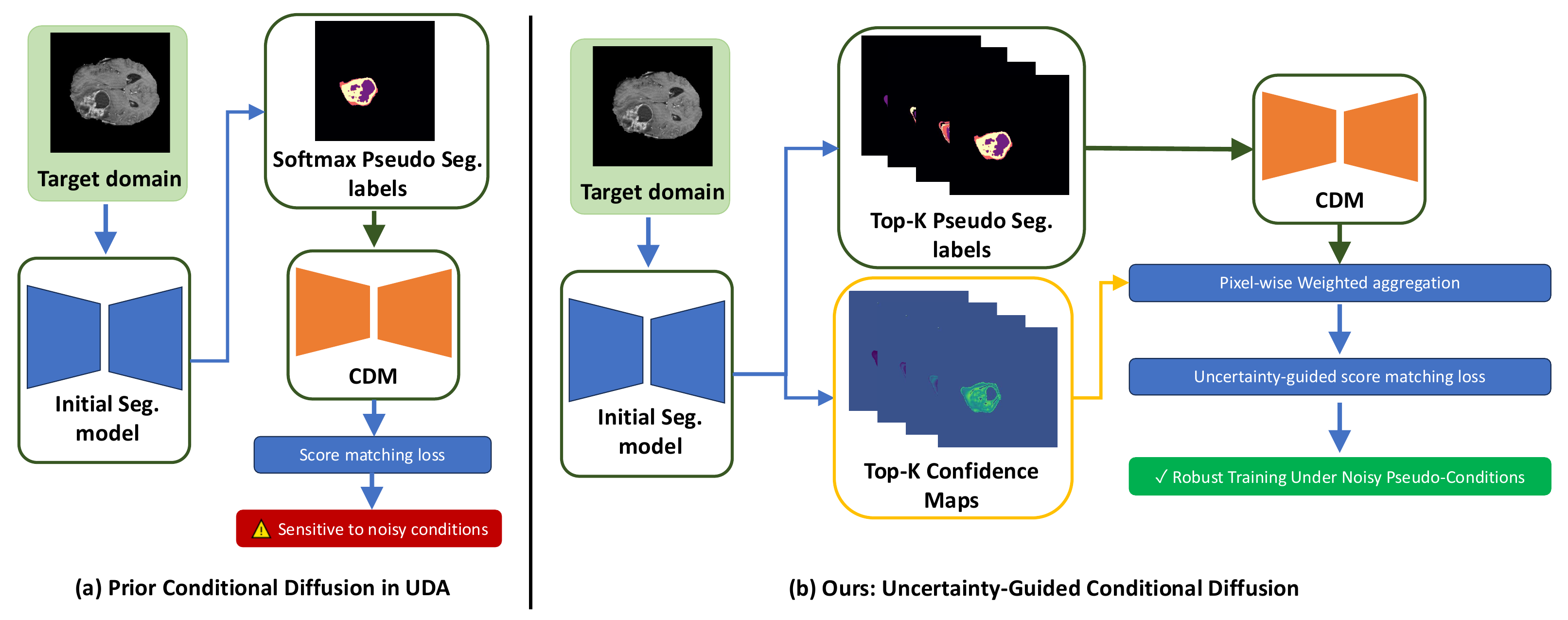}


\caption{The comparison between (a) Prior conditional diffusion in UDA and (b) our Uncertainty-guided conditional diffusion. 
}
\label{fig:compare}
\end{figure*}

Recently, diffusion models, particularly conditional diffusion models (CDMs), have been explored for UDA~\cite{ji2024diffusion,gong2024diffuse,luo2025noise}, mostly as generative augmentation modules. 
Masks are provided as conditional input for CDMs to generate corresponding syntheic images. These mask-image pairs are supposed to mimic target domain annotation-image pairs to adapt the model. The issue, however, is we do not have ground truth masks in the target domain in the first place. Some pipelines implicitly rely on stronger assumptions by training with (image, mask) pairs via style translation and reusing source masks as labels, treating source labels as directly transferable to target-style data~\cite{peng2023diffusion}.
Another realistic alternative is to use pseudo-labels, i.e., masks predicted by a source-domain-trained segmenter on target-domain images~\cite{shen2024controluda, gong2024diffuse, hu2024adaptdiff}. However, under domain shift, a source-domain-trained model's predictions on target images are often uncertain and error-prone; the noise will be amplified by CDM, yielding unrealistic synthetic mask-image pairs, and ultimately degrading model adaptation~\cite{luo2025noise}. See \cref{fig:compare} for illustrations.

We focus on this critical yet under-addressed challenge in UDA: \textbf{how to robustly train CDMs using noisy target-domain pseudo-labels}. An intuitive approach, borrowed from semi-supervised learning \cite{sohn2020fixmatch, li2023calibrating}, is to leverage the uncertainty of these labels by discarding high-uncertainty samples. However, given the significant domain gap, pseudo-labels are often excessively noisy. This causes vanilla uncertainty-based strategies to fail: it either discards too much data, leading to inefficient learning, or admits too much noise, ultimately derailing the training process.


In this paper, we propose a novel multi-channel uncertainty-guided CDM training for UDA. The key idea is to treat the segmentor output not as a single pseudo-label mask, but as a distribution over plausible labels. For each target domain image, the segmentation model produces a softmax probability vector at every pixel. Instead of only using the prediction (i.e., the Arg-Max mask), we argue that for uncertain prediction, model's confidence on all labels are valuable information and should be used.
During CDM training, we generate ranked pseudo-label maps (Arg-Max, Arg-2nd, Arg-3rd, etc.) as separate channels. \footnote{Here “map” denotes a full spatial label map (an arg-$k$ label at every pixel/voxel), not a single one-hot class channel.}
We use all these different channels to train the CDMs. Each channel has its own conditional score estimation. Aggregating these multi-channel score estimates weighted by their confidence results in a much more effective uncertainty-reweighted score-matching objective. This design has two benefits: (1) it preserves useful semantic guidance from high-confidence regions, and (2) it reduces the negative impact of incorrect pseudo-labels by downweighting uncertain predictions while still exploiting alternative plausible labels. An illustration is shown in \cref{fig:compare}. In practice, one important strength is that our method does not change the diffusion architecture; it only changes the training objective, making the implementation easy and robust.

We further provide theoretical insight into the proposed aggregation: under a squared-error criterion, weighting multiple conditional score estimates by the segmenter’s softmax probabilities yields the optimal convex combination with respect to a surrogate label distribution defined by the softmax outputs, highlighting why we should not collapse the softmax to an Arg-Max pseudo-label. 


To further strengthen the pseudo-conditions and uncertainty signals used by our diffusion training, we also introduce a supporting appearance-alignment component, Bézier adaptation. Rather than learning an unconstrained, free-form translation, Bézier adaptation applies a learnable Bézier-curve-based transformation with limited degrees of freedom and feature-guided optimization. This improves stability and generalization and, crucially for our main contribution, yields better initial segmentation predictions and more informative uncertainty maps in the target domain, improving uncertainty-guided diffusion training.

\begin{figure*}[t]
    \centering 
   \includegraphics[width=1\linewidth]{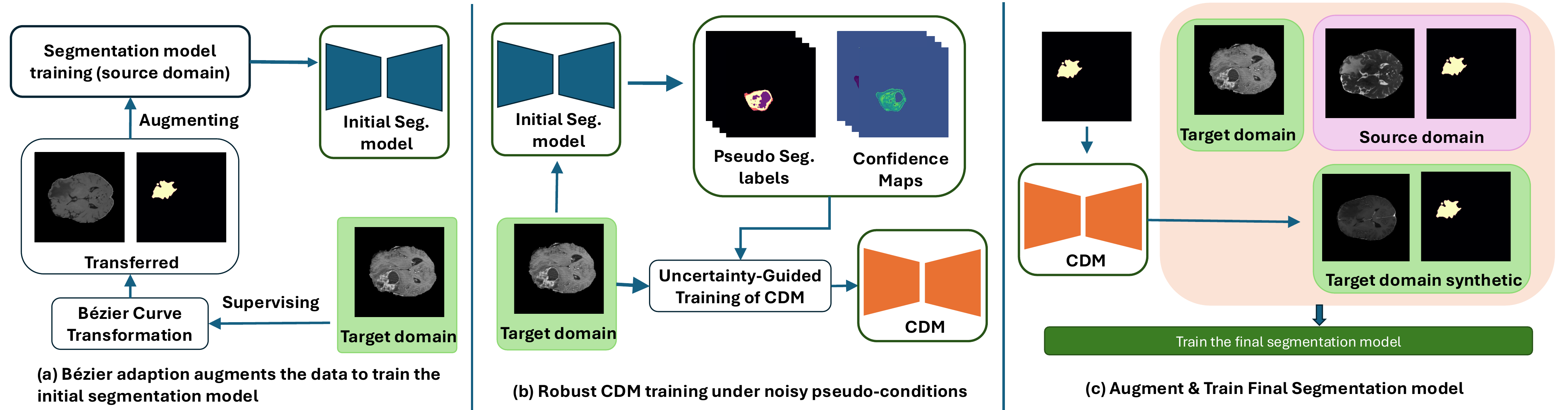}


\caption{The overview of our pipeline: Bézier adaptation, uncertainty guided CDM, and Plug-and-play integration with common UDA pipelines.
}
\label{fig:overall}
\end{figure*}

Putting these components together, we present UPDiff-UDA, a UDA framework that (i) enables robust target-domain CDM training under noisy pseudo-labels via uncertainty-guided score matching and (ii) produces stronger pseudo-conditions via stable, constrained appearance alignment. The trained CDM then generates labeled target-style image–mask pairs for augmentation, allowing us to train a high-quality target-domain segmentation model. Please see \cref{fig:overall} for an illustration. In summary, our main contributions are:

\begin{itemize}[topsep=0pt,itemsep=3pt,partopsep=0pt, parsep=0pt,leftmargin=.2in]
    \item \textbf{Multi-channel uncertainty-guided CDM training under noisy pseudo-label conditions}. We introduce a new score-matching objective that leverages the segmentation model’s ranked softmax predictions and pixel-wise confidence weighting to robustly train conditional diffusion in the target domain without ground-truth target masks. 

\item \textbf{Theoretical justification and robustness analysis.} We provide theoretical justification for the proposed confidence-weighted aggregation by establishing a minimum-MSE convex aggregation principle under an assumed per-pixel label distribution.
 
 
    \item \textbf{Bézier adaptation as a supporting pillar}. We propose a constrained, feature-guided Bézier-curve transformation that reduces appearance gaps and improves the quality of pseudo-labels/uncertainty estimates feeding into uncertainty-guided diffusion training.


\end{itemize}

\noindent
Using the robustly trained CDM, we synthesize labeled target-style image–mask pairs to strengthen target-domain training and improve segmentation performance. We demonstrate that our framework integrates with common UDA pipelines and consistently improves performance across datasets and modality shifts.

\section{Related Work}
\textbf{Unsupervised domain adaptation} (UDA) for segmentation has been extensively studied as a cost-effective solution to the challenge of acquiring high-quality segmentation annotations~\cite{hoffman2018cycada, kim2020learning, li2019bidirectional, tsai2018learning, zhao2024unsupervised, zhang2021prototypical, hu2025learn2synth}, particularly in the context of medical imaging, where manual annotations demand significant expertise and resources~\cite{chen2020unsupervised, dorent2020scribble, huo2018synseg, jiang2020psigan, liu2019susan, zhang2018translating}.
Self-training~\cite{bruggemann2023contrastive, hoyer2022daformer, xie2023sepico, wu2021one, shen2023loopda} is widely utilized in UDA due to the substantial performance gains achieved through pseudo-labeling. However, the gain of self-training is limited due to two aspects: First, the quality of pseudo-labels can be degraded due to the domain gap. Second, the available target domain images determine the performance upper bound of UDA methods. In contrast, our framework leverages the generative capacity of CDM, enabling us to generate an unlimited number of labeled target domain images, irrespective of the availability of target domain samples.

\noindent\textbf{GAN-based style transfer}~\cite{arjovsky2017wasserstein, goodfellow2014generative, karras2019style, mao2017least} mitigates domain gaps by adapting the image style of the source domain to match that of the target domain. This enables the segmentation model trained on style-transferred images to achieve improved performance in the target domain~\cite{hoffman2018cycada, jiang2020psigan, zhu2017unpaired}. However, the effectiveness of these domain style transfer methods is limited. Achieving a thorough and precise style transformation is highly challenging. In tumor regions, style transfer can be particularly problematic and may even have a counterproductive effect due to the sparsity and location variability of tumor regions. Although methods like CyCADA~\cite{hoffman2018cycada} use semantic consistency loss to force the translated image to have the same segmentation map as before, the need to use multiple loss functions impairs the model's ability to depict tumor regions. 

\noindent\textbf{Diffusion models}~\cite{dhariwal2021diffusion, ho2020denoising, nichol2021improved, songdenoising} are widely applied to vision tasks due to their ability to produce high-quality samples. For the segmentation task, diffusion models are applied to both natural images~\cite{amit2021segdiff, chen2023generalist,ji2023ddp, peng2023diffusion} and medical images~\cite{wu2024medsegdiff, wolleb2022diffusion, li2024spatial, xu2025topocellgen}. In terms of domain adaptation, \cite{peng2023diffusion} proposes a diffusion-based image translation framework guided by pixel-wise semantic labels for semantic segmentation. In Wang et al.~\cite{wang2023towards}, the diffusion model is utilized as an encoder to learn domain-invariant representations, facilitating UDA in medical image segmentation. To the best of our knowledge, we are among the first to leverage conditional diffusion models for directly augmenting the target domain with high-quality labeled synthetic images.

\section{Method}

Our \textit{UPDiff-UDA} framework is built around an \emph{uncertainty-guided conditional diffusion model (CDM)} for target-domain data augmentation (\cref{sec:cdm}), with a lightweight Bézier-curve-based style alignment module (\textit{Bézier adaptation}, \cref{sec:bezier}) as a supporting component.
The core challenge we address is that training a CDM in UDA requires conditional labels in the target domain, yet only \emph{noisy pseudo-labels} are available due to domain shift.
Naively conditioning a diffusion model on a single Arg-Max pseudo mask can propagate label errors into generation and undermine downstream segmentation.
To overcome this bottleneck, we propose a \emph{noise-robust training strategy} for conditional diffusion: instead of treating the segmenter output as one pseudo mask, we exploit the \emph{full softmax distribution} by constructing multiple ranked pseudo-label maps (Arg-Max, Arg-2nd, Arg-3rd, etc.) and using their pixel-wise confidence to reweight conditional score estimates during diffusion training.
This uncertainty-guided score matching objective enables the CDM to leverage useful semantic cues while suppressing the impact of incorrect pseudo conditions, resulting in more reliable target-style labeled synthesis.

As shown in \cref{fig:overall}(b), we train the CDM on target-domain images and then use it as a label-preserving generator.
Under the standard UDA assumption that the semantic label distribution is shared across domains, the CDM takes a segmentation mask (e.g., from the source domain) as condition and generates a corresponding target-style image, producing labeled synthetic pairs for augmentation.
These generated image--mask pairs can be combined with unlabeled target images and plugged into existing UDA pipelines to learn a strong target-domain segmentation model.
To further improve the quality of pseudo conditions and the associated uncertainty signals used for CDM training, we introduce \textit{Bézier adaptation} as a supporting alignment step: it applies a constrained, learnable Bézier-curve intensity transformation optimized via feature-space similarity, narrowing the appearance gap and yielding more reliable segmentation predictions on target images.
The overall framework is illustrated in \cref{fig:overall}.

\begin{figure*}[t]
    \centering 
   \includegraphics[width=1\linewidth]{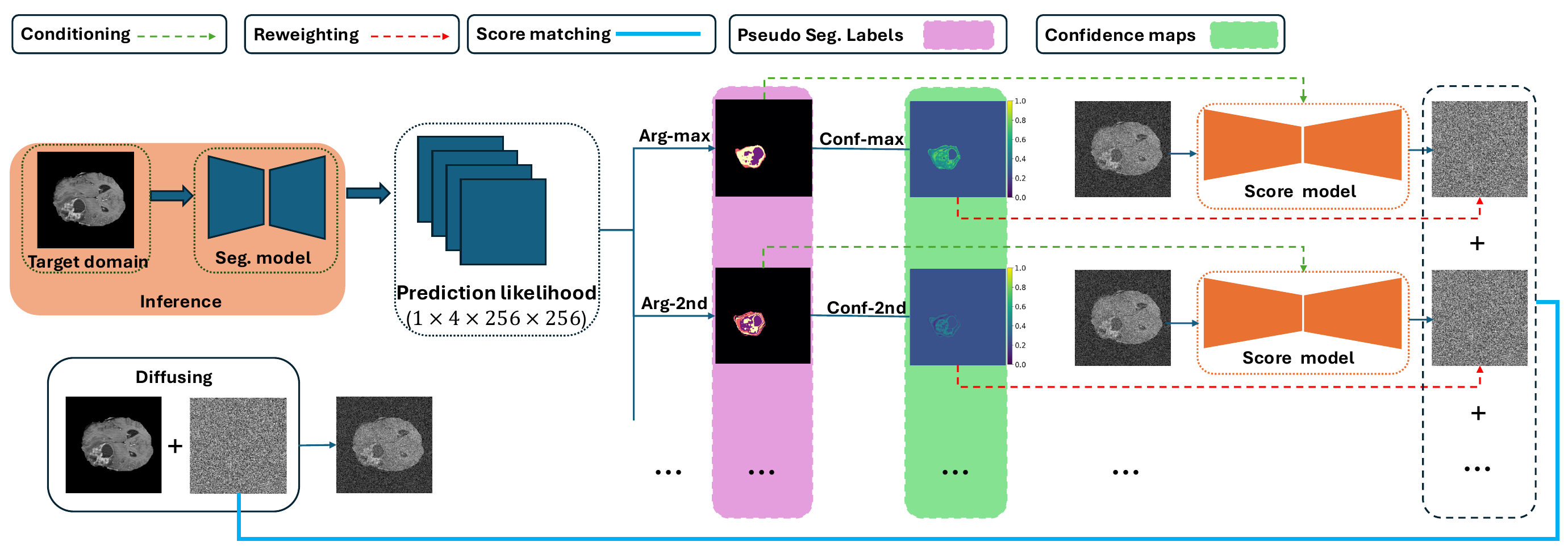}


\caption{An overview of the proposed uncertainty-guided CDM training framework. 
}
\label{fig:second}
\end{figure*}

\subsection{Preliminaries: score-based diffusion models} 
\textbf{Diffusion models} synthesize data by transforming Gaussian noise into samples via a reverse-time stochastic process.
Let $x_t\in\mathbb{R}^d$ denote the data state at time $t\in[0,T]$.
The forward diffusion process is described by the SDE
\begin{equation}
    dx_t = f(x_t,t)\,dt + g(t)\,dw_t,
    \label{eq:fp}
\end{equation}
where $f(\cdot,\cdot)$ is the drift, $g(\cdot)$ the diffusion coefficient, and $w_t$ is a standard Wiener process.
Let $p_t(x)$ be the marginal density of $x_t$.
Under mild regularity conditions, the corresponding reverse-time SDE is
\begin{equation}
    dx_t = \Big[f(x_t,t) - g(t)^2 \nabla_{x_t}\log p_t(x_t)\Big]dt + g(t)\,d\bar{w}_t,
    \label{eq:rp}
\end{equation}
where $\bar{w}_t$ is a reverse-time Wiener process.
Since the score $\nabla_{x_t}\log p_t(x_t)$ is intractable, it is approximated by a neural network $s_\theta$ trained via denoising score matching (DSM)~\cite{ho2020denoising}:
\begin{equation}
\mathcal{L}_{\mathrm{DSM}}(\theta)
=
\mathbb{E}_{t}\!\left[
\lambda(t)\,
\mathbb{E}_{(x_0,y)\sim p_0(X,Y)}\,
\mathbb{E}_{x_t\sim p_{t|0}(\cdot\mid x_0)}\,
\Big\|
s_\theta(x_t,y,t)
-
\nabla_{x_t}\log p_{t|0}(x_t\mid x_0)
\Big\|_2^2
\right],
\label{eq:dsm}
\end{equation}
where $p_{t|0}(x_t\mid x_0)$ is the Gaussian perturbation kernel implied by \cref{eq:fp}, and $\lambda(t)$ is a weighting schedule.

\subsection{Uncertainty-guided conditional diffusion model on the target domain} \label{sec:cdm}
We aim to train a target-domain conditional diffusion model (CDM) using only target images $X_t$ and an imperfect segmentation model $f_p$ (trained on the source domain and thus potentially unreliable on $X_t$).
We adapt $f_p$ to target images and use its predictions as conditional inputs for CDM training.
Unlike standard pseudo-labeling that uses only the Arg-Max prediction, we jointly utilize multiple semantic predictions from $f_p$.
For a given target image $x_0$, we generate multiple pseudo segmentation masks (Arg-Max, Arg-2nd, Arg-3rd, etc.), and use their corresponding confidence maps to aggregate multiple conditional score estimates in a pixel-wise manner. The overall training pipeline and an illustrative example of these ranked pseudo maps are shown in \cref{fig:second}.

\paragraph{Pseudo-label distribution and confidence maps.}
Let $p_u(k):=\mathrm{softmax}(f_p(x_0))_{u,k}$ be the segmentation probability for class $k$ at pixel/voxel $u$.
For each pixel, we view the softmax vector $\{p_u(k)\}_{k=1}^{|\mathcal{K}|}$ as a surrogate distribution over plausible labels in the target domain.
When the Arg-Max confidence is low, the Arg-2nd (or other ranked) class can have a substantial chance to be correct; thus incorporating the alternatives can mitigate pseudo-label errors in uncertain regions.

\paragraph{Uncertainty-reweighted score estimation.}
We feed pseudo-label maps derived from $f_p$ to the conditional score network $s_\theta$ to obtain label-conditional score estimates.
These estimates are aggregated using pixel-wise confidence weights to form a single uncertainty-reweighted score:
\begin{equation}
\hat{s}_{\theta}(x_t,t)
=
\sum_{k = 1}^{|\mathcal{K}|}
p(\cdot=k)\ \odot\ s_\theta (x_t, y=k, t),
\label{eq:agg_score_full}
\end{equation}
where $p(\cdot=k)$ denotes the per-pixel weight map $u\mapsto p_u(k)$ and ``$\odot$'' denotes the Hadamard (pixel-wise) product.

\paragraph{Uncertainty-guided score matching objective.}
We train the CDM by replacing the score network in \cref{eq:dsm} with the uncertainty-reweighted score $\hat{s}_\theta$:
\begin{equation}
\mathcal{L}_{\mathrm{UGSM}}(\theta)
=
\mathbb{E}_{t}\!\left[
\lambda(t)\,
\mathbb{E}_{x_0\sim p_t(X_t)}\,
\mathbb{E}_{x_t\sim p_{t|0}(\cdot\mid x_0)}\,
\Big\|
\hat{s}_\theta(x_t,t)
-
\nabla_{x_t}\log p_{t|0}(x_t\mid x_0)
\Big\|_2^2
\right].
\label{eq:ugsm}
\end{equation}
Optimizing \cref{eq:ugsm} ensures that high-confidence predictions contribute strong semantic guidance to the score estimation, while lower-confidence predictions are still utilized but downweighted, thereby reducing the negative impact of erroneous pseudo-labels.
Importantly, the reweighting is pixel-wise, reflecting the fact that confidence can vary across spatial locations.

\section{Theoretical justification: why multi-channel}
Our confidence-weighted aggregation is motivated by a minimum-MSE convex-combination principle: given a per-pixel label distribution, the squared-error optimal convex weights for aggregating label-conditional scores match that distribution. 
We instantiate this distribution using the segmenter softmax outputs, which clarifies why we need not collapse the softmax to an Arg-Max pseudo-label (Arg-Max is optimal only when predictions are nearly deterministic). 
We formalize this principle in \cref{thm:full_mse}.

\begin{theorem}[Minimum-MSE convex aggregation under an assumed label distribution]
\label{thm:full_mse}
Fix a diffusion time $t$ and pixel $u$.
Let $Y(u)\in\{1,\dots,|\mathcal{K}|\}$ be a discrete label variable with an \emph{assumed} per-pixel distribution
$q_u(k):=\mathbb{P}(Y(u)=k\mid x_0)$ (e.g., provided by a segmentation model).
For each label $\ell$, denote the label-conditional score at $u$ by
$S_\ell(u):=\nabla_{x_t(u)}\log p_t(x_t\mid Y=\ell)$.
Consider deterministic aggregated estimators of the form
\[
a_w(u)=\sum_{\ell=1}^{|\mathcal{K}|} w_u(\ell)\,S_{\ell}(u),
\qquad w_u\in\Delta^{|\mathcal{K}|-1}.
\]
Then the conditional mean squared error
\[
\mathbb{E}\!\left[
\left\|
a_w(u) - S_{Y(u)}(u)
\right\|_2^2
\,\middle|\,
x_0,x_t
\right]
\]
is minimized by $w_u=q_u$.
\end{theorem}

\noindent\textbf{Implication.}
Arg-Max conditioning corresponds to a one-hot $q_u$ and is optimal only when the label distribution is (nearly) deterministic; otherwise, probability-weighted aggregation yields the minimum-MSE convex combination under $q_u$.

\begin{proof}[compact]
Condition on $(x_0,x_t)$ and define $m_q(u):=\sum_{\ell} q_u(\ell)\,S_\ell(u)$.
For any fixed (deterministic) $a_w(u)$,
\[
\mathbb{E}\!\left[\|a_w(u)-S_{Y(u)}(u)\|_2^2\mid x_0,x_t\right]
=
\|a_w(u)-m_q(u)\|_2^2
+
C(u),
\]
where
\[
C(u)=\sum_{\ell} q_u(\ell)\|S_\ell(u)\|_2^2-\|m_q(u)\|_2^2
\]
is independent of $w_u$.
Thus the minimizer satisfies $a_w(u)=m_q(u)$, which is achieved by $w_u=q_u$.
\qed
\end{proof}

\noindent\textbf{Discussion.}
\cref{thm:full_mse} provides a minimum-MSE convex aggregation principle: the optimal weights equal the underlying per-pixel label distribution $q_u$. In our method, we instantiate $q_u$ using the segmenter softmax output $p_u(k)=\mathrm{softmax}(f_p(x_0))_{u,k}$, treating it as a surrogate estimate of target-domain label uncertainty under domain shift. Since this surrogate may be imperfect, the effectiveness of the aggregation depends on how well $p_u$ reflects the underlying label uncertainty; we analyze the impact of probability mismatch on the estimation error in the supplementary material.

\paragraph{Top-$K$ approximation (efficiency).}
Computing \cref{eq:agg_score_full} over all $|\mathcal{K}|$ labels can be expensive. In practice, we restrict aggregation to the top-$K$ classes per pixel and renormalize the weights. Let $\pi_u(1),\dots,\pi_u(K)$ be the indices of the $K$ largest values of $\{p_u(k)\}_{k=1}^{|\mathcal{K}|}$ and define $\tilde{y}^{(k)}(u):=\pi_u(k)$ and
\begin{equation}
\bar{c}^{(k)}(u)
:= \frac{p_u(\tilde{y}^{(k)}(u))}{\sum_{j=1}^{K} p_u(\tilde{y}^{(j)}(u)) + \epsilon},
\quad \sum_{k=1}^{K}\bar{c}^{(k)}(u)\approx 1.
\label{eq:conf_norm}
\end{equation}
We then approximate \cref{eq:agg_score_full} as
\begin{equation}
\hat{s}_{\theta}(x_t,t)
\approx
\sum_{k=1}^{K} \bar{c}^{(k)} \odot s_\theta(x_t,\tilde{y}^{(k)},t),
\label{eq:agg_score}
\end{equation}
where $K$ trades off computation and approximation fidelity (see supplementary material for further discussion).

\begin{figure*}[btp]
    \centering 
   \includegraphics[width=1\linewidth]{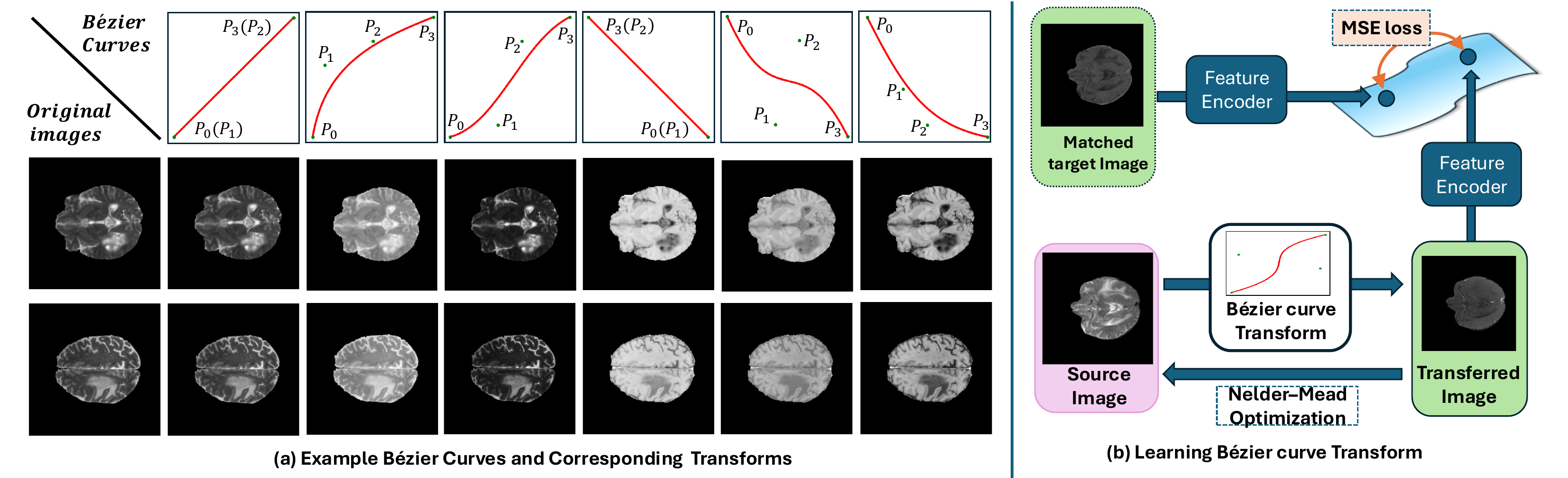}


\caption{The overview of Bézier-curve-based style transfer. (a) illustrates the impact of varying control points on Bézier curves and their effect on medical image styles. (b) is the control points optimization process of Bézier adaptation. 
}
\label{fig:bezier}
\end{figure*}

\paragraph{High-confidence collapse with per-pixel renormalization.}
We observed that in pixels where the segmenter is extremely confident, higher-rank predictions (Arg-2nd/Arg-3rd/$\dots$) are not informative (their probabilities are near zero) and may unnecessarily complicate uncertainty-guided CDM training. To handle this case, we apply a pixel-wise collapse rule during CDM training. Let $c^{(1)}(i,j)$ denote the Arg-Max confidence at pixel $(i,j)$ and let $\delta$ be a confidence threshold. For any $k>1$, if $c^{(1)}(i,j)>\delta$, we overwrite the higher-rank pseudo-labels with the Arg-Max label,
\[
\tilde{y}^{(k)}(i,j)\leftarrow \tilde{y}^{(1)}(i,j),
\]
so that alternative labels are discarded at these highly confident locations. We then set the corresponding confidence values to a uniform constant before per-pixel renormalization,
\[
c^{(k)}(i,j)\leftarrow \frac{1}{|C|},\quad k>1,
\]
where $|C|$ is the number of classes. After renormalization, the aggregation remains equivalent to conditioning on Arg-Max only, because all channels share the same label at $(i,j)$ (hence they produce the same conditional score estimate), while in pixels with $c^{(1)}(i,j)\le\delta$ we keep the original ranked pseudo-labels so that plausible alternatives can contribute in uncertain regions.

\subsection{Bézier adaptation} \label{sec:bezier}
Bézier adaptation is a lightweight intensity alignment module used to reduce the appearance gap before training our uncertainty-guided CDM. We model the cross-domain intensity mapping with a cubic Bézier curve, which provides a smooth nonlinear transformation controlled by only a few parameters. Compared with GAN-based translation, this constrained parameterization is more stable and less likely to introduce structural artifacts; compared with histogram matching, it is flexible enough to fit a specific source–target domain pair.

To learn the transformation, we optimize the Bézier control points so that transformed source images become closer to target images. Because pixel-wise comparison is unreliable under large modality shifts, we measure similarity in a deep feature space extracted by a pretrained encoder. Concretely, we first select representative source prototypes via 
K-means in feature space, retrieve their nearest neighbors from the target set, and then fit Bézier parameters by minimizing the feature-space MSE between each matched pair. Since the optimization landscape can be noisy and the parameter space is low-dimensional, we use a derivative-free optimizer (Nelder–Mead) for stable fitting. The resulting Bézier transformation produces better-aligned images, which in turn improves the quality of pseudo-labels and uncertainty maps used by our uncertainty-guided diffusion training. The complete Bézier adaptation pipeline is in \cref{fig:bezier}(b). More details about Bézier adaptation are in the supplementary material.

\section{Experiment}
\label{sec_experiment_details}
In the experimental section, we mainly evaluate the effectiveness of our method on unsupervised domain adaptation (UDA) for medical image segmentation.

\noindent\textbf{Datasets.}
We use three benchmark datasets (i.e., BraTS 2023 dataset~\cite{kazerooni2024brain}, Multi-Modality Whole Heart Segmentation dataset~\cite{zhuang2016multi}, and Abdominal Multi-Organ datasets~\cite{kavur2021chaos, landman2015miccai}) to show the effectiveness of our method. Please refer to the supplementary material for more details about the datasets. 





\noindent\textbf{Dataset preprocessing}
For the \textbf{BraTS} and \textbf{Multi-Organ} datasets, each 3D image is normalized using min-max scaling, with voxel intensities rescaled to the range $[0,1]$. For the MM-WHS dataset, we use the preprocessed images provided by Wang et al.~\cite{wang2023towards}.

\noindent\textbf{Baselines.} 
`No adaptation' denotes the baseline results obtained by evaluating a segmentation model trained solely on source domain data without applying any UDA techniques. CycleGAN~\cite{zhu2017unpaired}, CyCADA~\cite{hoffman2018cycada}, SIFA~\cite{chen2020unsupervised}, and PSIGAN~\cite{jiang2020psigan} are representative GAN-based UDA methods. ADVENT~\cite{vu2019advent} enhances UDA performance by introducing entropy minimization losses, while FDA~\cite{yang2020fda} leverages the Fourier Transform to reduce domain discrepancies. GenericSSL~\cite{wang2023towards} uses diffusion models to extract domain-invariant representations. FPL+~\cite{wu2024fpl+} employs cross-domain data augmentation and dual-domain pseudo label generation to effectively mitigate domain shift. Diff-style~\cite{peng2023diffusion} employs a classifier-guided diffusion model for image style transfer, effectively mitigating domain gaps.

\noindent\textbf{Implementation details.}
We use a U-Net with ResNet34 as the backbone for the segmentation task. The input images are augmented with random rotation, random scale, and Bézier-curve-based style augmentation.  For the selected UDA method, we enhance performance by adding our generated target domain images, paired with their corresponding segmentation conditions, into the labeled training set, while also applying Bézier adaptation as an additional augmentation during training. \textbf{Ours-AD}, \textbf{Ours-GS}, and \textbf{Ours-PL} denote the integration of our proposed generated data with ADVENT~\cite{vu2019advent}, GenericSSL~\cite{wang2023towards}, and pseudo-labeling (details in the supplementary material), respectively. 


\begin{figure}[t]
    \centering 
   \includegraphics[width=1.01\linewidth]{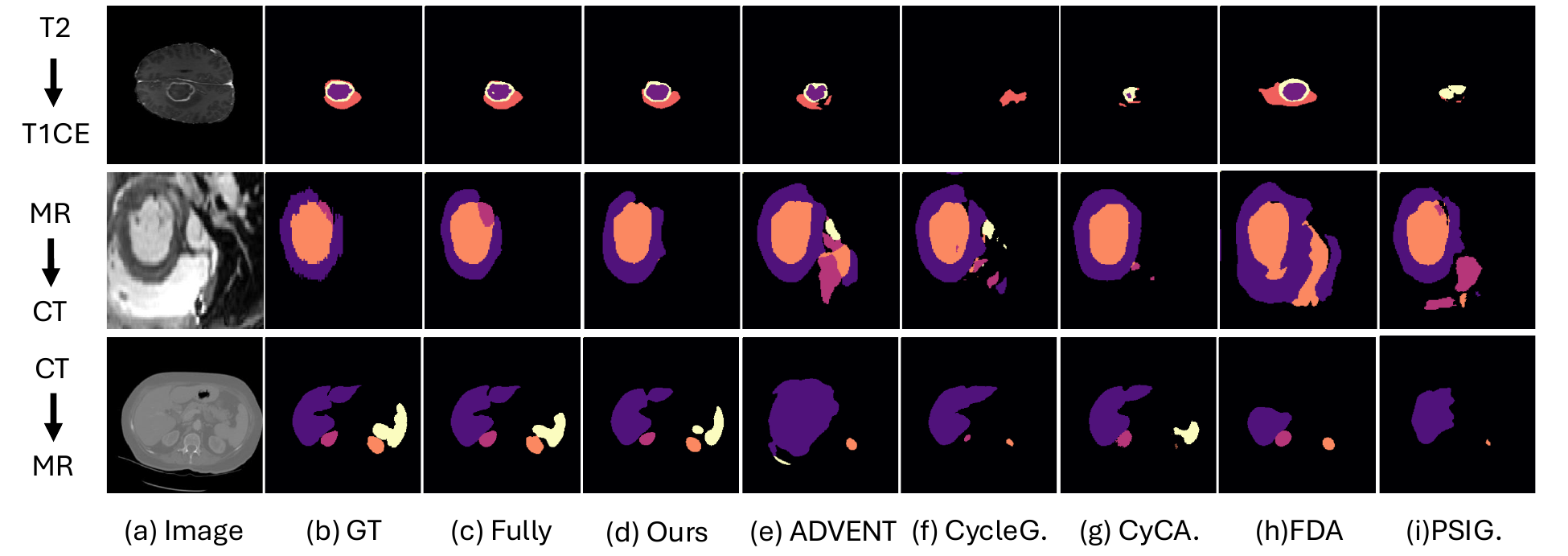}


\caption{Segmentation results of Ours-PL and other baselines. From top to bottom: (1) BraTS dataset – Purple, orange, and yellow represent NCR, ED, and ET classes, respectively; (2) MM-WHS dataset – Purple, pink, orange, and yellow represent LVM, LAB, LVB, and AA classes, respectively; (3) Multi-Organ dataset – Purple, pink, orange, and yellow represent Liver, R. Kidney, L. Kidney, and Spleen classes, respectively.
}
\label{qua_res_seg}
\end{figure}

\begin{figure}[t]
    \centering 
   \includegraphics[width=0.9\linewidth]{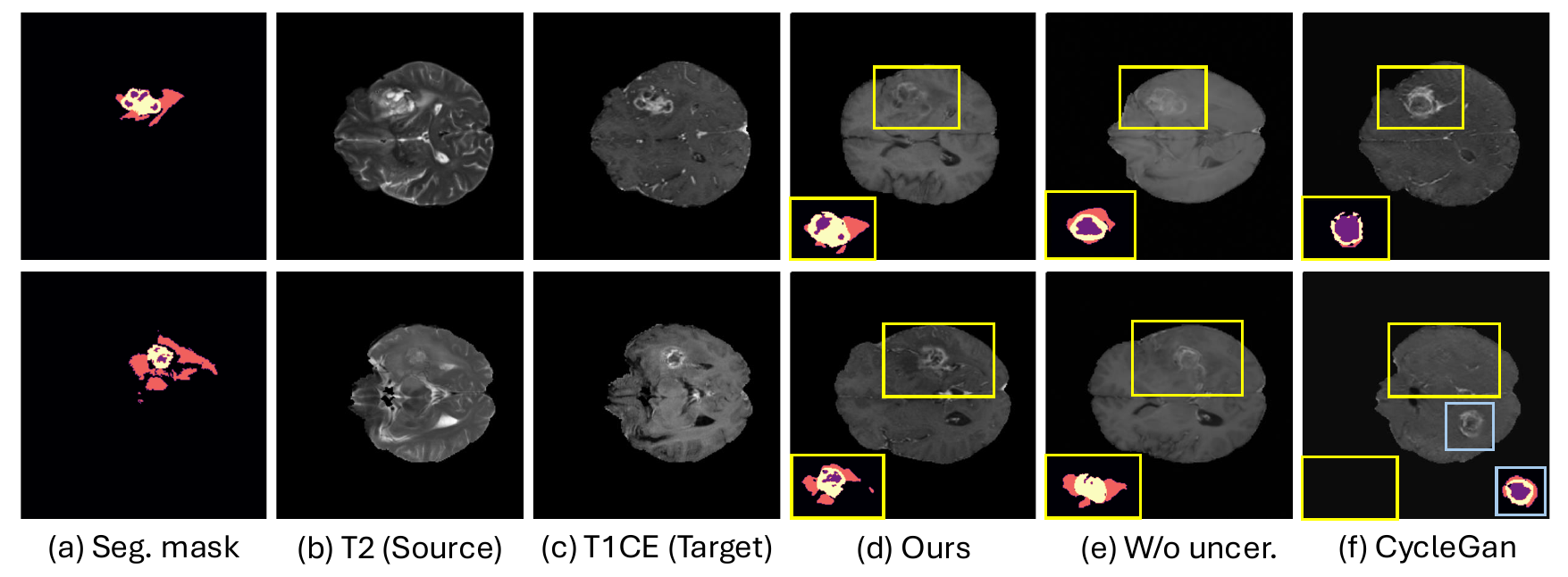}


\caption{Qualitative results for generative models. We also show the segmentation results of the tumor region. Our method (d) is much better than baselines: CDM trained without uncertainty (e), and CycleGAN (f). In the second row, CycleGAN even hallucinates the tumor at the wrong location.
}
\label{qua_res_gen}
\end{figure}
\subsection{Results}
\label{sec:results}
We compare our UPDiff-UDA augmented UDA methods with various baselines on three benchmark datasets. We report Dice (Dice) as the primary metric for the main comparisons in the paper; 95\% Hausdorff distance (95HD) and additional results are provided in the supplementary material. In the ablation study, we report Dice and 95HD to comprehensively analyze the impact of each component.

\textbf{BraTS} is a challenging UDA dataset due to the variability of tumors across different modalities. As shown in \cref{table1}, our method outperforms other baselines under the UDA setting. When T1 is the target domain, our \textit{UPDiff-UDA} approach significantly augments the ADVENT method, improving the Dice score by 17.0\%. When adapting T2 domain images to T1CE, \textbf{Ours-PL} outperforms other methods by a large margin. The generative-based baselines perform worse on this dataset because they rely on style transfer, which struggles in tumor regions due to the variability and sparsity of tumors.

\textbf{MM-WHS} serves as a strong benchmark for UDA methods due to the realistic domain gap between CT and MRI scans, particularly in contrast, noise characteristics, and anatomical appearance. Our method achieves impressive performance. As shown in \cref{table2}, our \textit{UPDiff-UDA} approach boosts ADVENT~\cite{vu2019advent} substantially, achieving a 19.7\% improvement in Dice score. 

\begin{table*}[t]
\centering
\caption{Comparison of different methods on the BraTS dataset (Dice \% $\uparrow$).}
\tiny{
\setlength{\tabcolsep}{8pt}{
\begin{tabular}{c|cccc|cccc}
    \toprule[0.8pt]
    Target domain & \multicolumn{4}{c|}{T1} & \multicolumn{4}{c}{T1CE} \\
    \midrule[0.5pt]
    \multirow{2}{*}{Method} 
    & WT & TC & ET & Ave. 
    & WT & TC & ET & Ave. \\
    \midrule[0.5pt]
    No Adaptation & 8.30 & 8.81 & 5.34 & 7.48 
                  & 9.96 & 20.35 & 14.57 & 14.96 \\
    Fully supervised & 73.30 & 61.63 & 41.63 & 58.86  
                     & 75.70 & 83.79 & 80.37 & 79.95 \\
    \midrule[0.5pt]
    CycleGAN & 7.21 & 5.95 & 3.88 & 5.68 
             & 12.55 & 23.93 & 12.49 & 16.32 \\
    CyCADA & 13.00 & 18.25 & 12.08 & 14.44 
           & 22.26 & 35.71 & 16.79 & 24.92 \\
    ADVENT & 54.24 & 49.49 & 32.91 & 45.55  
           & 40.67 & 48.28 & 34.81 & 41.25 \\
    FDA & 32.52 & 36.21 & 22.87 & 30.53  
        & 39.91 & 53.17 & 35.77 & 42.95 \\
    SIFA & 55.43 & 50.72 & 33.81 & 46.66 
         & 42.37 & 49.74 & 35.62 & 42.58 \\
    PSIGAN & 47.62 & 40.92 & 25.57 & 38.04 
           & 21.68 & 32.83 & 17.98 & 24.16 \\
    GenericSSL & 57.99 & 53.62 & 32.93 & 48.18 
               & 46.07 & 55.18 & 35.30 & 45.52 \\
    FPL+ & 58.97 & 54.83 & 34.68 & 49.49 
         & 48.72 & 56.17 & 36.63 & 47.17 \\
    Diff-style & 46.11 & 46.91 & 31.25 & 41.42 
               & 37.69 & 45.12 & 19.18 & 34.00 \\
    \midrule[0.5pt]
    \textbf{Ours-AD} & \underline{63.53} & \bf 57.71 & \bf 38.69 & \bf 53.31  
                     & \underline{57.99} & 61.77 & 34.38 & 51.38 \\
    \textbf{Ours-GS} & \bf 65.09 & 55.55 & 38.03 & \underline{52.89} 
                     & 46.34 & \underline{67.35} & \bf 44.33 & \underline{52.67} \\
    \textbf{Ours-PL} & 60.90 & \underline{56.95} & \underline{38.21} & 52.02  
                     & \bf 62.47 & \bf 67.89 & \underline{44.12} & \bf 58.16 \\
    \bottomrule[0.8pt]
\end{tabular}
}
}
\label{table1}
\end{table*}

\begin{table*}[t]
\centering
\caption{Comparison of different methods on the MM-WHS dataset (Dice \% $\uparrow$).}
\tiny{
\setlength{\tabcolsep}{5pt}{
\begin{tabular}{c|ccccc|ccccc}
    \toprule[0.8pt]
    & \multicolumn{5}{c|}{Cardiac MRI $\rightarrow$ Cardiac CT} 
    & \multicolumn{5}{c}{Cardiac CT $\rightarrow$ Cardiac MRI}\\
    \midrule[0.5pt]
    \multirow{2}{*}{Method} 
    & \multicolumn{5}{c|}{Dice (\%) $\uparrow$} 
    & \multicolumn{5}{c}{Dice (\%) $\uparrow$} \\
    \cmidrule[0.5pt]{2-11}
    & LVM & LAB & LVB & AA & Ave. 
    & LVM & LAB & LVB & AA & Ave. \\
    \midrule[0.5pt]
    No Adaptation & 63.72 & 66.73 & 82.84 & 68.19 & 70.37 
                  & 25.94 & 22.11 & 45.25 & 10.31 & 25.90 \\
    Fully supervised & 88.33 & 89.69 & 94.70 & 83.10 & 88.95 
                     & 90.94 & 94.47 & 92.98 & 96.13 & 93.63 \\
    \midrule[0.5pt]
    CycleGAN & 74.76 & 83.02 & 88.66 & 84.34 & 82.70 
             & 65.91 & 36.21 & 79.55 & 18.84 & 50.13 \\
    CyCADA & 69.78 & 87.63 & 84.69 & 65.43 & 76.88 
           & 64.56 & 47.64 & 81.55 & 31.39 & 56.28 \\
    ADVENT & 75.08 & 79.19 & 89.11 & 87.90 & 82.82 
           & 62.60 & 47.63 & 80.92 & 33.15 & 56.08 \\
    FDA & 59.09 & 33.88 & 73.90 & 17.99 & 46.22 
        & 35.82 & 14.77 & 53.69 & 0.74 & 26.26 \\
    SIFA & 65.72 & 78.36 & 81.59 & 71.82 & 74.37 
         & 51.57 & \underline{67.15} & 78.23 & \bf 66.07 & 65.75 \\
    PSIGAN & 13.90 & 22.26 & 27.42 & 63.24 & 31.71 
           & \underline{69.95} & 42.81 & 79.04 & 39.79 & 57.90 \\
    GenericSSL & \bf 87.95 & \underline{91.88} & \underline{91.02} & \bf 91.06 & \bf 90.48 
               & \bf 74.16 & 55.07 & 86.13 & 54.55 & 67.48 \\
    FPL+ & 75.50 & 81.20 & 86.40 & 78.30 & 80.35 
         & 59.84 & 54.97 & 79.12 & 44.37 & 59.58 \\
    Diff-style & 51.71 & 36.92 & 31.83 & 64.77 & 46.31 
               & 57.30 & 25.89 & 64.69 & 3.12 & 37.75 \\
    \midrule[0.5pt]
    \textbf{Ours-AD} & 79.04 & 85.23 & 89.45 & 85.92 & 84.91 
                     & 68.31 & 63.97 & \bf 86.77 & 56.59 & \underline{68.91} \\
    \textbf{Ours-GS} & \underline{86.84} & \bf 92.04 & \bf 92.63 & 89.07 & \underline{90.14} 
                     & 63.10 & \bf 70.56 & \underline{86.24} & \underline{61.66} & \bf 70.39 \\
    \textbf{Ours-PL} & 79.18 & 84.50 & 86.78 & \underline{89.23} & 84.92 
                     & 66.95 & 63.37 & 83.83 & 50.00 & 66.04 \\
    \bottomrule[0.8pt]
\end{tabular}
}
}
\label{table2}
\end{table*}

\begin{table*}[ht]
\centering
\caption{Comparing on the Abdominal Multi-Organ dataset (Dice \% $\uparrow$).}
\tiny{
\setlength{\tabcolsep}{5pt}{
\begin{tabular}{c|ccccc|ccccc}
    \toprule[0.8pt]
    & \multicolumn{5}{c|}{Abdominal MRI $\rightarrow$ Abdominal CT} 
    & \multicolumn{5}{c}{Abdominal CT $\rightarrow$ Abdominal MRI}\\
    \midrule[0.5pt]
    \multirow{2}{*}{Method} 
    & Liver & R. Kid & L. Kid & Spleen & Avg. 
    & Liver & R. Kid & L. Kid & Spleen & Avg. \\
    \midrule[0.5pt]
    No Adaptation & 9.66 & 70.32 & 61.45 & 17.05 & 39.62 
                  & 22.12 & 57.94 & 59.52 & 56.35 & 48.98 \\
    Fully supervised & 89.04 & 85.49 & 90.50 & 87.22 & 88.06 
                     & 83.60 & 83.69 & 77.01 & 65.90 & 77.55 \\
    \midrule[0.5pt]
    CycleGAN & 74.09 & 67.88 & 62.03 & 76.24 & 70.06 
             & 57.15 & 51.62 & 55.38 & 56.59 & 55.18 \\
    CyCADA & 63.31 & 69.60 & 72.06 & 74.09 & 69.77 
           & 53.59 & 60.18 & 60.74 & 58.70 & 58.30 \\
    ADVENT & 21.18 & 67.98 & 72.19 & 14.95 & 44.08 
           & 20.77 & 57.94 & 59.52 & 56.35 & 48.64 \\
    FDA  & 71.55 & 81.27 & 71.02 & 80.06 & 75.97 
         & 51.16 & 64.20 & 63.90 & 67.96 & 61.81 \\
    SIFA & 41.34 & 37.92 & 32.96 & 53.59 & 41.45 
         & 48.43 & 46.55 & 58.63 & 56.27 & 52.47 \\
    PSIGAN & 63.53 & 76.94 & \underline{77.09} & 71.01 & 72.14 
           & 34.20 & 51.65 & 53.99 & 58.79 & 49.66 \\
    GenericSSL & 36.79 & 56.33 & 36.22 & 9.72 & 34.77  
               & \underline{62.87} & 53.03 & 45.21 & 49.99 & 52.78 \\
    FPL+ & 74.20 & 79.15 & 71.60 & 78.45 & 75.85  
         & 55.90 & 66.85 & 64.30 & 68.10 & 63.79 \\
    Diff-style & 41.52 & 73.38 & 76.85 & 76.23 & 67.00 
               & 29.15 & 57.94 & 58.84 & 57.44 & 50.84 \\
    \midrule[0.5pt]
    \textbf{Ours-AD} & \underline{81.03} & 84.50 & 75.30 & \underline{88.67} & \underline{82.38} 
                     & 54.56 & 67.84 & 69.80 & \underline{75.34} & 66.89 \\
    \textbf{Ours-GS} & 75.05 & \bf 86.72 & 77.03 & 84.24 & 80.76 
                     & \bf 68.73 & \underline{72.61} & \bf 75.37 & 73.25 & \bf 72.49 \\
    \textbf{Ours-PL} & \bf 84.89 & \underline{85.74} & \bf 82.33 & \bf 90.47 & \bf 85.86 
                     & 59.02 & \bf 74.83 & \underline{70.78} & \bf 79.44 & \underline{71.02} \\
    \bottomrule[0.8pt]
\end{tabular}
}
}
\label{table3}
\end{table*}

The results of \textbf{Multi-Organ} are shown in \cref{table3}. Although ADVENT~\cite{vu2019advent} achieves subpar results on this dataset, our \textit{UPDiff-UDA} pipeline significantly improves its performance, surpassing other baselines. By combining our \textit{UPDiff-UDA} with the pseudo-labeling framework, the \textbf{Ours-PL} surpasses other baselines by a large margin. GenericSSL~\cite{wang2023towards} underperforms on this dataset, while the incorporation of our synthetic target domain images significantly improves its performance, highlighting the effectiveness of our pipeline.

Overall, the performance gains achieved by our \textit{UPDiff-UDA} demonstrate the effectiveness of the proposed method. Qualitative segmentation results are presented in \cref{qua_res_seg}. The Qualitative synthetic results in \cref{qua_res_gen} highlight the ability of our uncertainty-guided CDM to produce high-fidelity target domain images with accurate alignment to the conditional masks.

\begin{table}[ht]
\begin{minipage}[t]{0.48\textwidth}
\caption{Ablation study of components.}
\label{abla}
\resizebox{\linewidth}{!}{ 
\setlength{\tabcolsep}{1mm}{
\begin{tabular}{c|cccc|cccc}
    \toprule[0.8pt]
    Target domain & \multicolumn{8}{c}{T1C} \\
    \midrule[0.5pt]
    \multirow{2}{*}{Method} & \multicolumn{4}{c|}{Dice (\%) $\uparrow$} & \multicolumn{4}{c}{95HD $\downarrow$} \\
    \cmidrule[0.5pt]{2-9}
    & WT & TC & ET & Average & WT & TC & ET & Average  \\
    \midrule[0.5pt]
    w/o real & 54.43 & 60.33 & 33.75 & 49.51 & 25.83 & 39.10 & 41.38 & 35.44 \\
    Mean-teacher & 60.20 & 62.08 & 35.36 & 52.55 & 36.86 & 51.89 & 45.17 & 44.64 \\
    w/o threshold & 57.30 & 58.13 & 30.54 & 48.66 & 32.42 & 37.81 & 44.37 & 38.20 \\
     w/o uncertainty & 59.05 & 62.94 & 36.66 & 52.88 & 23.65 & 39.30 & 43.73 & 35.56\\
    \textbf{Ours-PL} & \bf 62.47 & \bf 67.89 & \bf 44.12 & \bf 58.16 & \bf 20.89  & \bf 23.07  & \bf 27.84 & \bf 23.93 \\
    \bottomrule[0.8pt]
\end{tabular}
}
}
\end{minipage}
\hfill
\begin{minipage}[t]{0.5\textwidth}
\caption{Ablation study of cold start.}
\label{abla_un}
\resizebox{\linewidth}{!}{ 
\setlength{\tabcolsep}{1mm}{
\begin{tabular}{c|cc|ccccc}
    \toprule[0.8pt]
    Target domain & \multicolumn{7}{c}{T1C} \\
    \midrule[0.5pt]
    \multirow{2}{*}{Method} & \multicolumn{2}{c|}{Accuracy ($\times 10^2$)} & \multicolumn{5}{c}{Uncertainty evaluation ($\times 10^3$)} \\
    \cmidrule[0.5pt]{2-8}
    & Dice$\uparrow$ & 95HD$\downarrow$ & AURC$\downarrow$ & E-AURC$\downarrow$ & ECE$\downarrow$& NLL$\downarrow$& Brier$\downarrow$  \\
    \midrule[0.5pt]
    No adaptation & 14.96 & 61.23 & 0.86 & 0.78 & 7.92 & 41.27 & 7.64 \\
    Histogram matching & 11.98 & 64.72 & 0.78 & 0.65 & 9.61 & 62.36 & 9.78  \\
    Fourier transform & 14.72 & 58.23 & 0.39 & 0.34 & 6.23 & 46.11 & 6.39\\
    Random Bézier & 41.88 & 34.87 & 0.06 & 0.05 & 2.72 & 13.65 & 3.22 \\
    \textbf{Bézier adapt}  & \bf 48.45 & \bf 30.25 & \bf 0.05 & \bf 0.04 & \bf 2.34 & \bf 11.50 & \bf 2.79 \\
    \bottomrule[0.8pt]
\end{tabular}
}
}
\end{minipage}
\end{table}

\subsection{Ablation Study}
\label{sec:ablation}
In this section, we conduct extensive ablation studies to justify the effectiveness of different components. More ablation study results about the impact of varying hyperparameters are in the supplementary material. 

\noindent\textbf{Components.} 
We conduct experiments on the BraTS 2023 dataset to validate the effectiveness of each component of our method, with a specific focus on domain adaptation from T2 to T1CE and the mean-teacher-like pseudo-labeling framework. The results are shown in \cref{abla}. In ``w/o uncertainty'', the CDM is trained by pseudo-labels obtained by applying the argmax function to the initial segmentation model output. Without our proposed training strategy, a significant performance drop is observed in target domain segmentation. ``mean-teacher'' denotes the result when the pseudo-labeling framework is applied without augmentation from our high-quality target image generations. It reflects the effectiveness of our strategy in boosting the pseudo-labeling method. ``w/o real'' shows the result of the segmentation model trained with our proposed generations without real source domain images. We observe that even without the support of real images and their ground truth, the segmentation model trained on our labeled synthetic target images achieves respectable performance in target domain segmentation. This demonstrates the high quality of our generation.

\noindent\textbf{Number of confidence maps.}
We conduct an ablation study on the number of confidence maps ($k$) employed for uncertainty-guided CDM training.
As illustrated in \cref{num_unc_maps}, the better performance is achieved when $k=2$.
In practice, we also observe that, except for Arg-Max and
Arg-2nd, other predictions contain less information (i.e., most of their confidence maps exhibit near-zero values). In the T2$\rightarrow$T1CE setting, we evaluate voxel-level Top-$K$ accuracy, defined as whether the ground-truth class is included among the $K$ most probable predicted classes. The Top-1/2/3/4 accuracies are 0.9783, 0.9954, 0.9987, and 1.0000, respectively. These results indicate that the Top-2 predictions already contain sufficient useful information, whereas the Arg-3rd and Arg-4th maps contribute only marginal additional gains. In practice, larger values of $K$ mainly lead to higher VRAM usage and smaller feasible batch sizes. 
On two NVIDIA H100 80GB GPUs~\footnote{For all other results, the diffusion model is trained on three NVIDIA RTX 8000 GPUs. The two NVIDIA H100 80GB GPUs are used only in this analysis to estimate the feasible batch sizes for different values of $K$ in the Top-$K$ setting.}, the maximum batch sizes for $K=1$, $K=2$, and $K=3$ are 48, 24, and 16, respectively. As a result, increasing $K$ substantially slows training convergence.
We may safely drop these predictions in order to save GPU memory and maximize our batch size,
given limited computational resources. 

\noindent\textbf{Preliminary adaptation.}
In this paper, we employ Bézier adaptation to improve the quality of pseudo-labels and their associated confidence maps, ensuring they are sufficiently reliable to guide our strategy for training a conditional diffusion model (CDM) capable of generating high-quality target-domain images. To validate its effectiveness, we compare Bézier adaptation with alternative approaches using metrics from two perspectives: (i) the quality of confidence maps and (ii) the final UDA performance. For confidence map evaluation, we adopt metrics from the uncertainty estimation literature. The area under the risk-coverage curve (AURC) quantifies failure prediction, where a lower value indicates that correctly predicted samples receive higher confidence scores, enabling more effective filtering. Excess-AURC (E-AURC) serves as the normalized variant of AURC~\cite{moon2020confidence, li2023confidence}. Calibration assesses the alignment between model confidence and the true likelihood of correctness. To this end, we use expected calibration error (ECE)~\cite{naeini2015obtaining}, the Brier score~\cite{brier1950verification}, and negative log-likelihood (NLL) to evaluate calibration quality.

As shown in \cref{abla_un}, compared with other style transfer baselines, our Bézier adaptation consistently achieves better performance on both segmentation accuracy and uncertainty-aware metrics. These gains indicate that Bézier adaptation yields more trustworthy pseudo-labels and better-calibrated confidence maps from the initial segmenter, which directly strengthens the supervision signal for our uncertainty-guided CDM training. We further evaluate the impact of different style transfer strategies on end-to-end UDA performance. As reported in \cref{style_transfer_prior}, Bézier adaptation again provides the strongest results among all alternatives, highlighting its effectiveness for mitigating appearance gaps and supporting it as a key component of our overall framework.

\begin{table}[t]
\centering
\begin{minipage}[t]{0.49\linewidth}
\centering
\caption{Ablation study on the impact of different style transfer methods.}
\label{style_transfer_prior}
\tiny
\resizebox{\linewidth}{!}{%
\begin{tabular}{c|cccc|cccc}
    \toprule[0.8pt]
    Target domain & \multicolumn{8}{c}{T1C} \\
    \midrule[0.5pt]
    \multirow{2}{*}{Style transfer} & \multicolumn{4}{c|}{Dice (\%) $\uparrow$} & \multicolumn{4}{c}{95HD $\downarrow$} \\
    \cmidrule[0.5pt]{2-9}
    & WT & TC & ET & Average & WT & TC & ET & Average  \\
    \midrule[0.5pt]
    Source only & 26.88 & 25.71 & 13.88 & 22.16 & 56.21 & 98.27 & 105.10 & 86.53 \\
    CycleGAN  & 41.09 & 42.73 & 25.54 & 36.45 & 34.42 & 51.94 & 59.74 & 48.70 \\
    CyCADA & 57.25 & 58.15 & 30.89 & 48.76 & 29.15 & 33.90 & 38.92 & 33.99 \\
    PSIGAN & 34.76 & 41.97 & 19.26 & 32.00 & 31.30 & 74.24 & 75.92 & 60.49 \\
    FDA & 41.81 & 45.24 & 29.12 & 38.72 & 39.01 & 61.39 & 66.79 & 55.73 \\
    \bf Ours-B\'ezier & \bf 62.47 & \bf 67.89 & \bf 44.12 & \bf 58.16 & \bf 20.89 & \bf 23.07 & \bf 27.84 & \bf 23.93 \\
    \bottomrule[0.8pt]
\end{tabular}%
}
\end{minipage}\hfill
\begin{minipage}[t]{0.49\linewidth}
\centering
\caption{Ablation study on the $\#$ of confidence maps used in UPDiff-UDA.}
\label{num_unc_maps}
\tiny
\resizebox{\linewidth}{!}{%
\begin{tabular}{c|cccc|cccc}
    \toprule[0.8pt]
    Target domain & \multicolumn{8}{c}{T1C} \\
    \midrule[0.5pt]
    \multirow{2}{*}{$k$} 
    & \multicolumn{4}{c|}{Dice (\%) $\uparrow$} 
    & \multicolumn{4}{c}{95HD $\downarrow$} \\
    \cmidrule[0.5pt]{2-9}
    & WT & TC & ET & Average 
    & WT & TC & ET & Average  \\
    \midrule[0.5pt]
    1 & 59.05 & 62.94 & 36.66 & 52.88 
      & 23.65 & 39.30 & 43.73 & 35.56 \\
    \textbf{2} & \bf 62.47 & \bf 67.89 & \bf 44.12 & \bf 58.16 
               & \textbf{20.89} & \bf 23.07 & \bf 27.84 & \bf 23.93 \\
    \bottomrule[0.8pt]
\end{tabular}%
}
\end{minipage}
\end{table}

\section{Conclusion}
We propose \textit{UPDiff-UDA}, a unified UDA framework for medical image segmentation that enables robust \emph{target-domain conditional diffusion training} under noisy pseudo-labels. Our core contribution is an uncertainty-guided score matching objective that leverages ranked softmax pseudo-label maps and pixel-wise confidence weighting, with theoretical support from a minimum-MSE convex-aggregation principle under a surrogate label distribution. We further introduce a feature-guided Bézier-curve adaptation module for constrained appearance alignment to improve pseudo-conditions and uncertainty estimates. Together, our method delivers consistent improvements across UDA benchmarks, while plug-and-play with existing pipelines.

\section*{Acknowledgments}
This work was partially supported by grants NSF CCF-2144901, NIH R01NS143143, and R01CA297843.
%
%


\bibliographystyle{splncs04}
\bibliography{main}

 




\clearpage
\title{Multi-Channel Uncertainty-Weighted Score Matching for Conditional Diffusion in Medical UDA \\--- Supplementary Material ---} 

\titlerunning{UPDiff-UDA}

\author{Chen Li\thanks{Email: Chen Li (li.chen.8@stonybrook.edu).}\inst{1} \and
Meilong Xu\inst{1}\ \and
Xiaoling Hu\inst{2} \and 
Weimin Lyu\inst{1} \and
Chao Chen\inst{1} }


\institute{Stony Brook University, Stony Brook, NY, USA \and
Massachusetts General Hospital and Harvard Medical School, MA, USA}

\maketitle
\setcounter{section}{6}
\setcounter{table}{7}
\setcounter{figure}{6}
\setcounter{equation}{7}
\setcounter{theorem}{1}

\section*{Overview}
This supplementary document provides additional material for the paper titled ``Multi-Channel Uncertainty-Weighted Score Matching for Conditional Diffusion in Medical UDA.'' It includes extended experimental results, implementation details, pseudocode, and additional theoretical analyses that further support the main claims and findings of the paper.

\section{Effect of Surrogate Distribution Mismatch}
\begin{theorem}[Excess MSE due to probability mismatch (ideal scores)]
\label{thm:mismatch_excess_supple}
Fix diffusion time $t$ and pixel $u$.
Let $Y^\star(u)\in\{1,\dots,|\mathcal{K}|\}$ be the unknown true label at $u$ with true label distribution
$r_u(k)=\mathbb{P}(Y^\star(u)=k\mid x_0)$.
Let $q_u(k)$ be a surrogate distribution used for aggregation (e.g., the segmenter softmax, or its top-$K$ truncation and renormalization).
For each label $k$, denote the ideal label-conditional score at $u$ by
$S_k(u):=\nabla_{x_t(u)}\log p_t(x_t\mid Y=k)$.
Define the mixture scores
\[
m_r(u):=\sum_{k=1}^{|\mathcal{K}|} r_u(k)\,S_k(u),
\qquad
m_q(u):=\sum_{k=1}^{|\mathcal{K}|} q_u(k)\,S_k(u).
\]
Consider deterministic aggregated estimators $a_w(u)=\sum_k w_u(k)S_k(u)$ with $w_u\in\Delta^{|\mathcal{K}|-1}$ and define the conditional risk
\[
\mathcal{R}(w_u):=
\mathbb{E}\!\left[\|a_{w}(u)-S_{Y^\star(u)}(u)\|_2^2\mid x_0,x_t\right].
\]
Then the excess risk incurred by using $q_u$ instead of the Bayes-optimal weights satisfies
\begin{equation}
\mathcal{R}(q_u)-\min_{w_u\in\Delta^{|\mathcal{K}|-1}}\mathcal{R}(w_u)
=
\|m_q(u)-m_r(u)\|_2^2.
\label{eq:excess_gap_supple}
\end{equation}
\end{theorem}

\begin{proof}
Condition on $(x_0,x_t)$; note that $r_u$ depends only on $x_0$ and is therefore fixed under this conditioning.
Let $Z:=S_{Y^\star(u)}(u)$ so that $\mathbb{E}[Z\mid x_0,x_t]=m_r(u)$.
For any fixed (deterministic) vector $a$,
\[
\mathbb{E}\big[\|a-Z\|_2^2\mid x_0,x_t\big]
=
\|a-m_r(u)\|_2^2
+
\underbrace{\Big(\mathbb{E}[\|Z\|_2^2\mid x_0,x_t]-\|m_r(u)\|_2^2\Big)}_{\mathrm{const}},
\]
where $\mathrm{const}$ does not depend on $a$.
Applying this identity to $a=a_w(u)$ shows the minimum is achieved at $a=m_r(u)$, i.e., $w_u=r_u$.
Therefore $\min_{w_u}\mathcal{R}(w_u)=\mathcal{R}(r_u)$ and
\[
\mathcal{R}(q_u)-\min_{w_u}\mathcal{R}(w_u)
=
\|a_q(u)-m_r(u)\|_2^2
=
\|m_q(u)-m_r(u)\|_2^2,
\]
proving \cref{eq:excess_gap_supple}.
\qed
\end{proof}
\noindent\textbf{Interpretation.}
\cref{thm:mismatch_excess_supple} shows that using a surrogate label distribution $q_u$ (e.g., the segmenter softmax) instead of the true distribution $r_u$ incurs an \emph{exact} excess conditional MSE equal to $\|m_q(u)-m_r(u)\|_2^2$, where $m_r$ and $m_q$ are the corresponding probability-weighted mixtures of label-conditional scores. Thus, probability mismatch introduces a bias in the aggregated score, and the performance degradation is directly determined by how far the surrogate-induced mixture score deviates from the Bayes-optimal mixture.

\section{Effect of Top-$K$ Truncation}
\begin{lemma}[Irreducible gap due to top-$K$ truncation]
\label{lem:topk_gap_supple}
Fix diffusion time $t$ and pixel $u$.
Let $r_u(k)=\mathbb{P}(Y^\star(u)=k\mid x_0)$ be the true label distribution and
$S_k(u)=\nabla_{x_t(u)}\log p_t(x_t\mid Y=k)$ the ideal label-conditional score.
Let $A_u\subset\{1,\dots,|\mathcal{K}|\}$ be the set of top-$K$ labels selected by a surrogate distribution (e.g., segmenter softmax).
Define the in-set mass and missing mass under the true distribution:
\[
\alpha(u) := \sum_{k\in A_u} r_u(k),\qquad \delta(u):=1-\alpha(u).
\]
Let the full Bayes mixture score be $m_r(u):=\sum_{k} r_u(k)S_k(u)$, and define the best achievable
top-$K$ mixture score (optimal within convex combinations supported on $A_u$):
\begin{equation}
m_{r,A}(u)\in \arg\min_{\substack{w\in\Delta^{K-1}}}
\left\|\sum_{k\in A_u} w(k)S_k(u)-m_r(u)\right\|_2^2.
    \label{eq:proj_def_supple}
\end{equation}
Then the minimum achievable conditional MSE using only labels in $A_u$ satisfies
\begin{align}
\min_{w\in\Delta^{K-1}}
\mathbb{E}\!\left[
\left\|\sum_{k\in A_u} w(k)S_k(u)-S_{Y^\star(u)}(u)\right\|_2^2
\,\middle|\, x_0,x_t
\right]
&=
\mathrm{Var}\!\left(S_{Y^\star(u)}(u)\,\middle|\,x_0,x_t\right)
\nonumber\\
&\quad+
\|m_{r,A}(u)-m_r(u)\|_2^2.
\label{eq:topk_exact_gap_supple}
\end{align}
Moreover, if $\|S_k(u)\|_2\le M(u)$ for all $k$, then
\begin{align}
\|m_{r,A}(u)-m_r(u)\|_2
&\le 2M(u)\,\delta(u), \label{eq:topk_gap_bound_supple}\\
\Rightarrow\quad
\text{excess MSE due to top-$K$}
&\le 4M(u)^2\,\delta(u)^2. \nonumber
\end{align}
In particular, when the true label has nonzero probability outside $A_u$ (i.e., $\delta(u)>0$),
there is an unavoidable approximation gap that vanishes as $\delta(u)\to 0$ (e.g., for larger $K$).
\end{lemma}

\begin{proof}
Condition on $(x_0,x_t)$ and let $Z:=S_{Y^\star(u)}(u)$ with mean $m_r(u)=\mathbb{E}[Z\mid x_0,x_t]$.
For any deterministic vector $a$,
\[
\mathbb{E}[\|a-Z\|_2^2\mid x_0,x_t]=\|a-m_r(u)\|_2^2+\mathrm{Var}(Z\mid x_0,x_t),
\]
so minimizing over $a$ constrained to the convex hull $\mathrm{conv}\{S_k(u)\}_{k\in A_u}$ is equivalent to minimizing
$\|a-m_r(u)\|_2^2$ over that convex hull. This is achieved by the Euclidean projection of $m_r(u)$ onto
$\mathrm{conv}\{S_k(u)\}_{k\in A_u}$, which is equivalently represented by some convex weights $w\in\Delta^{K-1}$.
By definition \cref{eq:proj_def_supple}, this minimizer equals $m_{r,A}(u)$, yielding \cref{eq:topk_exact_gap_supple}.

For the bound, write
\[
m_r(u)=\sum_{k\in A_u} r_u(k)S_k(u) + \sum_{k\notin A_u} r_u(k)S_k(u).
\]
Let $\bar{m}_{r,A}(u):=\sum_{k\in A_u}\frac{r_u(k)}{\alpha(u)}S_k(u)$ (for $\alpha(u)>0$).
Then $\|\bar{m}_{r,A}(u)\|_2\le M(u)$ since it is a convex combination of $\{S_k(u)\}$, and
\[
m_r(u)=\alpha(u)\bar{m}_{r,A}(u)+\sum_{k\notin A_u} r_u(k)S_k(u).
\]
Hence,
\begin{align}
\|m_r(u)-\bar{m}_{r,A}(u)\|_2
&=
\left\|(1-\alpha(u))\bar{m}_{r,A}(u)
+\sum_{k\notin A_u} r_u(k)S_k(u)\right\|_2 \nonumber\\
&\le (1-\alpha(u))\|\bar{m}_{r,A}(u)\|_2
+\sum_{k\notin A_u} r_u(k)\|S_k(u)\|_2 \nonumber\\
&\le \delta(u)M(u)+\delta(u)M(u)
=2M(u)\delta(u).
\end{align}
Finally, since $m_{r,A}(u)$ is the minimizer of $\|a-m_r(u)\|_2$ over $a\in\mathrm{conv}\{S_k(u)\}_{k\in A_u}$,
we have $\|m_{r,A}(u)-m_r(u)\|_2 \le \|\bar{m}_{r,A}(u)-m_r(u)\|_2$, which implies \cref{eq:topk_gap_bound_supple}.
\qed
\end{proof}


\noindent\textbf{Interpretation.}
\cref{lem:topk_gap_supple} highlights an inherent limitation of restricting aggregation to a top-$K$ label set $A_u$.
Even with perfect label-conditional scores $\{S_k(u)\}$, a top-$K$ estimator is confined to the convex hull of $\{S_k(u)\}_{k\in A_u}$ and therefore cannot represent contributions from labels outside $A_u$.
Accordingly, the lemma decomposes the minimum achievable conditional MSE into (i) an unavoidable variance term
$\mathrm{Var}\!\left(S_{Y^\star(u)}(u)\mid x_0,x_t\right)$ and (ii) an additional squared bias term
$\|m_{r,A}(u)-m_r(u)\|_2^2$ arising solely from truncation.
This bias term is the \emph{irreducible gap} caused by missing probability mass outside $A_u$; it is strictly positive whenever
$\delta(u)=1-\sum_{k\in A_u} r_u(k)>0$.
Moreover, \cref{eq:topk_gap_bound_supple} shows that the truncation-induced error is controlled by the missing mass $\delta(u)$ and vanishes as $\delta(u)\to 0$ (e.g., as $K$ increases or when the surrogate selection captures most of the true mass).
In practice, we find that using $K=2$ already captures the dominant uncertainty in most regions and yields strong performance while remaining computationally efficient.

\section{Implementation Details}

We train the segmentation model for 200 epochs with a batch size of 64. The segmentation network is optimized by Adam with learning rate of 1e-4. The diffusion model is trained for 250,000 iterations with a batch size of 7, using three NVIDIA RTX 8000 GPUs, 40-core Intel(R) Xeon(R) Gold 6248 CPUs @ 2.50GHz, and 754 GB of RAM.

The pretrained feature extractor used in~\cref{alg:bead} is adopted from the autoencoder in Stable-Diffusion-v1-4.

In Tab.~4, ``w/o threshold'' is the ablation result without using the thresholding strategy described in the main text from line 195 to line 205, which uses Arg-Max in high confidence regions as the pseudo-labels for CDM training. This result reflects the effectiveness of this well-designed strategy.

\section{Pseudo Code}
In this part, we provide the pseudo code for uncertainty-guided score matching in \cref{alg:udsm} and Bézier adaptation in \cref{alg:bead}.

\begin{figure*}[h]
\centering
\scriptsize

\begin{algorithm}[H]
\caption{Training with uncertainty guided score matching}\label{alg:udsm}
\KwIn{Target image set $X_t$, pseudo-labeling network $f_p$, transition kernel $p_{t|0}$, confidence threshold $\delta$ }
\KwOut{Conditional score network $s_\theta$}
\While{ not converged}{
Sample $x_0 (N\times N)$ from $X_t$, and $t$ from $[0, T]$\;
Sample $x_t$ from $p_{t|0}$\;
Compute pseudo-labels and uncertainties:
$c^{(k)}, \tilde{y}^{(k)} \leftarrow f_p(x_0), k\in C$\;

\For{$(i, j) \in \{0, \dots, N\} \times \{0, \dots, N\}$}{
\If{$ c_{ij}^{(1)}  > \delta $ }{
$\tilde{y}_{ij}^{(k)} = \tilde{y}_{ij}^{(1)}, c_{ij}^{(k)} = 1/|C|, \forall k \in C$\;
}
}

Update $\theta$ by gradient descent step on
$ \nabla_\theta \| \sum_{k=1}^{|C|} c^{(k)} s_\theta - g_t(x_0, x_t, y)\|_2^2$\;
}
\end{algorithm}

\vspace{0.5em}

\begin{algorithm}[H]
\caption{Bézier adaption}\label{alg:bead}
\KwIn{Source image set $X_s$, Target image set $X_t$, Pretrained image feature extractor $\phi$}
\KwOut{Optimized control point sets $P_1^*, P_2^*$}
Extract source image features $X_s^f = \phi(X_s)$\;
Extract target image features $X_t^f = \phi(X_t)$\;
Apply K-means clustering on $X_s^f$ with $n_s$ clusters to find cluster centers $\{c_1, c_2, \ldots, c_{n_s}\}$\;
\For{$j \gets 1$ \KwTo $n_s$}{
    Select the closest image (feature) $x_j \in X_s$ to cluster center $c_j$ as the prototype\;
    Find the closest image (feature) $x_p \in X_t$ to $x_j$ as the matched target image\;
    \textbf{Initialize} random control points and transfer source domain image with Bézier curve as $B(x_j)$\;
    Define the \textbf{objective function} as $f_l = \| \phi(B(x_j)) - \phi(x_p) \|_2^2$\;
    \textbf{Optimize} control points $P_1, P_2$ with the Nelder--Mead method\;
}
\end{algorithm}

\end{figure*}

\section{Bézier Adaptation Details}
\paragraph{Bézier Adaptation.}
\textbf{Bézier curves} provide an effective parameterization for intensity transformation because they (i) enable flexible nonlinear mappings between pixel-value distributions, (ii) are controlled by only a few parameters, and (iii) remain smooth and stable, which is desirable for medical images where preserving anatomical structure is critical.
We adopt a \emph{cubic} Bézier curve to map normalized source intensities to normalized target intensities, i.e., $[0,1]\to[0,1]$.

A Bézier curve of degree $n$ is a parametric curve defined by $n\!+\!1$ control points $\{P_i\}_{i=0}^{n}$ with $P_i=(x_i,y_i)$.
Using the Bernstein basis, the curve is
\begin{equation}
B(t)=\sum_{i=0}^{n}\binom{n}{i}(1-t)^{n-i}t^i P_i,\quad t\in[0,1],
\end{equation}
following~\cite{farin2002curves}.
In our cubic case ($n=3$), the curve is determined by four control points:
two endpoints $P_0$ and $P_3$ anchoring the mapping at $t=0$ and $t=1$, and two interior control points $P_1,P_2$ controlling the curvature.
The mapping function is obtained by evaluating the curve (typically with $x_0=0,x_3=1$ and a monotonicity constraint), yielding a smooth nonlinear intensity transform.
As a result, a wide range of contrast/brightness changes can be modeled with only a small number of parameters, as illustrated in Fig.4(a).

\paragraph{Learning a dataset-level transformation.}
We learn a set of Bézier control points that reduces the appearance gap between source and target domains.
Given a source image $x_s$, we first identify a target image $x_t$ that is most similar in \emph{feature space} and then optimize the Bézier curve so that the transformed image $\mathcal{T}_{\phi}(x_s)$ matches $x_t$ as closely as possible, where $\phi$ denotes the curve parameters (control points).
Comparing images in pixel space is often unreliable under domain shift; instead, we use a pretrained encoder $E(\cdot)$ to extract domain-invariant structural features and minimize a feature-level distance:
\begin{equation}
\min_{\phi}\ \big\|E(\mathcal{T}_{\phi}(x_s)) - E(x_t)\big\|_2^2.
\end{equation}

\paragraph{Prototype matching and optimization.}
Directly matching every source image to the target domain is computationally expensive and redundant.
In practice, we first select representative \emph{prototypes} from the source domain by running $K$-means clustering in the encoder feature space.
For each selected source prototype, we retrieve its nearest neighbor in the target feature space to form a matched source--target pair.
We then optimize the Bézier control points by minimizing the feature MSE over these matched pairs.
The full pipeline is shown in Fig.4(b), and pseudo code is provided in \cref{alg:bead}.

\paragraph{Why derivative-free optimization.}
Because matched source and target images are not perfectly aligned and can exhibit structural differences, the resulting objective can be noisy and may lead to unstable gradients.
Since the Bézier parameterization is low-dimensional, we adopt the Nelder--Mead method to optimize the control points.
Compared with unconstrained, deep-learning-based style transfer, Bézier adaptation produces smooth, intensity-preserving transformations that reduce the risk of corrupting anatomical structures.
Moreover, using multiple matched pairs introduces controlled diversity in the learned mappings, helping avoid overfitting to a small set of examples.

\section{Pseudo-labeling Details}
\label{pl_details_supple}
The pseudo-labeling framework used in this paper is mean-teacher-like. We have two models: the student model and the teacher model. The parameters of the teacher model are updated by the EMA method from the student model. The teacher model is for generating pseudo-labels and confidence maps for the training of the student model on the target domain. Considering the imperfection of pseudo-labels, we filter out the low-quality pseudo-labels by thresholding on confidence maps. Only the pseudo-labels with high enough confidence are kept and used for supervising the student model. 
During training, as the quality of pseudo-labels progressively improves, we gradually relax the threshold to incorporate more high-quality pseudo-labels into the learning process. This threshold decreases from $\tau_u$ to $\tau_l$ during training.
We supervise the training of the student model using both real labels and selected pseudo-labels, with a combination of cross-entropy and Dice losses. The loss contribution from pseudo-labels is scaled by a weighting factor $\lambda$.


\section{Dataset Details}
\textbf{The Brain Tumor Segmentation (BraTS)} focus on brain tumor sub-region segmentation. The sub-regions are composed by enhancing tumor (ET), the tumor core (TC), and the whole tumor (WT). There are multiple modalities available in the Brats dataset, i.e., T2, Flair, T1, and T1CE. In this paper, we consider the domain adaptation from T2 to the other 3 modalities. The regions of the GD-enhancing tumor (ET — label 3), the peritumoral edematous/invaded tissue (ED — label 2), and the necrotic tumor core (NCR — label 1) are annotated in images. The TC class comprises regions ET and NCR. The WT class comprises regions ET, ED, and NCR. The training set is constructed by 142600 images (71300 source domain images, 71300 target domain images), and the validation set is constructed by 12400 source domain images. The test set is constructed by 38905 target domain images. We generate 30514 target domain synthetic images using segmentation masks from the source domain. 

\noindent
\textbf{The Multi-Modality Whole Heart Segmentation (MM-WHS)} consist of 20 unpaired volumetric cardiac images, including CT and MRI scans. The annotations for the segmentation of four cardiac structures are provided: the left ventricle myocardium (LVM), left atrium blood cavity (LAB), left ventricle blood cavity (LVB), and ascending aorta (AA). The training set is constructed by 2304 source domain and 2016 target domain images. The validation set is constructed by 576 source domain images. The test set is constructed by 867 target domain images. For this dataset, we generate 2304 target domain synthetic images conditioning on masks from the source domain. 

\noindent
\textbf{The Abdominal Multi-Organ datasets} contain two modalities: MRI images from the ISBI 2019 CHAOS Challenge dataset~\cite{kavur2021chaos} and CT images from the Multi-Atlas Labeling Beyond the Cranial Vault - Workshop and Challenge~\cite{landman2015miccai}. Four abdominal organs are labeled for segmentation: liver, right kidney (R. Kid), left kidney (L. Kid), and spleen. For MRI to CT, the training set comprises 491 source domain and 2449 target domain images. The validation set comprises 30 source domain images. The test set comprises 1048 target domain images. For CT to MRI, the training set comprises 2449 source domain and 491 target domain images. The validation set comprises 282 source domain images. The test set comprises 126 target domain images. For CT-to-MRI adaptation, we generate 2,449 synthetic MRI images conditioned on source domain segmentation masks. Similarly, for MRI-to-CT adaptation, we generate 1,964 synthetic CT images using the same conditioning strategy.


\section{Supplementary 95HD Results}
In this section, we report the 95HD results on the BraTS, MM-WHS, and Multi-Organ datasets in \cref{table_1}, \cref{table_2}, and \cref{table_3}, respectively, complementing the corresponding Dice scores presented in Tables 1–3 of the main paper. Across all three benchmarks, our method consistently achieves strong performance under both Dice and 95HD, demonstrating that the proposed approach improves not only overlap accuracy but also boundary quality.

\begin{table*}[t]
\centering
\caption{Comparison of different methods on the BraTS dataset.}
\label{table_1}
\tiny{
\setlength{\tabcolsep}{1.8mm}{
\begin{tabular}{c|cccc|cccc}
    \toprule[0.8pt]
    Target domain & \multicolumn{4}{c|}{T1} & \multicolumn{4}{c}{T1CE} \\
    \midrule[0.5pt]
    \multirow{2}{*}{Method} & \multicolumn{4}{c|}{95HD $\downarrow$} & \multicolumn{4}{c}{95HD $\downarrow$}\\
    \cmidrule[0.5pt]{2-9}
    & WT & TC & ET & Average & WT & TC & ET & Average \\
    \midrule[0.5pt]
    No Adaptation & 60.90 & 71.04 & 114.37 & 82.10 & 59.10 & 59.55 & 65.03 & 61.23 \\
    Fully supervised & 11.74 & 23.12 & 29.49 & 21.45 & 10.85 & 8.74 & 15.23 & 11.61 \\
    \midrule[0.5pt]
    CycleGAN & 55.47 & 65.78 & 80.31 & 67.19 & 57.40 & 58.61 & 63.30 & 59.77 \\
    CyCADA & 58.84 & 112.97 & 129.56 & 100.45 & 61.17 & 76.48 & 80.84 & 72.83 \\
    ADVENT & 25.64 & 37.61 & 41.53 & 34.93 & 33.96 & 44.02 & 46.79 & 41.59 \\
    FDA & 73.01 & 98.52 & 104.61 & 92.05 & 39.20 & 56.11 & 59.39 & 51.57 \\
    SIFA & \underline{23.81} & 35.68 & 39.25 & 32.91 & 31.59 & 43.24 & 45.82 & 40.22 \\
    PSIGAN & 25.70 & 46.09 & 54.08 & 41.96 & 41.47 & 70.43 & 74.69 & 62.20 \\
    GenericSSL & 37.50 & 36.65 & 41.28 & 38.48 & 53.41 & 46.18 & 50.91 & 50.16 \\
    FPL+ & 33.41 & 34.86 & 38.05 & 35.44 & 49.06 & 45.37 & 48.22 & 47.55 \\
    Diff-style & 43.41 & 46.99 & 49.08 & 46.49 & 37.60 & 44.66 & 53.62 & 45.29 \\
    \midrule[0.5pt]
    \textbf{Ours-AD} & \bf 20.31 & \bf 28.03 & \bf 32.99 & \bf 27.11 & \bf 20.77 & 33.62 & 39.38 & \underline{31.26} \\
    \textbf{Ours-GS} & 41.26 & \underline{29.53} & \underline{35.64} & 35.48 & 57.20 & \bf 22.39 & \bf 25.61 & 35.07 \\
    \textbf{Ours-PL} & 24.97 & 34.57 & 37.94 & \underline{32.50} & \underline{20.89} & \underline{23.07} & \underline{27.84} & \bf 23.93 \\
    \bottomrule[0.8pt]
\end{tabular}
}
}
\end{table*}

\begin{table*}[t]
\centering
\caption{Comparison of different methods on the MM-WHS dataset.}
\label{table_2}
\tiny{
\setlength{\tabcolsep}{0.9mm}{
\begin{tabular}{c|ccccc|ccccc}
    \toprule[0.8pt]
    & \multicolumn{5}{c|}{Cardiac MRI $\rightarrow$ Cardiac CT} & \multicolumn{5}{c}{Cardiac CT $\rightarrow$ Cardiac MRI}\\
    \midrule[0.5pt]
    \multirow{2}{*}{Method} & \multicolumn{5}{c|}{95HD $\downarrow$} & \multicolumn{5}{c}{95HD $\downarrow$} \\
    \cmidrule[0.5pt]{2-11}
    & LVM & LAB & LVB & AA & Average & LVM & LAB & LVB & AA & Average \\
    \midrule[0.5pt]
    No Adaptation & 9.03 & 18.33 & 10.67 & 15.50 & 13.38 & 68.60 & 64.35 & 60.88 & 72.45 & 66.57 \\
    Fully supervised & 2.58 & 5.01 & 8.30 & 13.30 & 7.30 & 3.37 & 15.37 & 3.86 & 6.26 & 7.21 \\
    \midrule[0.5pt]
    CycleGAN & 14.78 & 38.68 & 6.44 & 36.64 & 24.14 & 47.04 & 49.12 & 47.22 & 48.39 & 47.94 \\
    CyCADA & 19.96 & 17.12 & 9.67 & 25.05 & 17.95 & 24.96 & 54.06 & \underline{32.02} & 61.37 & 43.10 \\
    ADVENT & 15.16 & 14.74 & 5.15 & \underline{13.74} & 12.19 & 43.36 & 54.07 & 37.89 & \underline{33.56} & 42.22\\
    FDA & 14.30 & 29.76 & 20.08 & 62.38 & 31.63 & 60.02 & 53.54 & 60.37 & 72.05 & 61.50 \\
    SIFA & 20.18 & 27.05 & 21.73 & 26.56 & 23.88 & 77.08 & 55.41 & 34.35 & 77.67 & 61.13\\
    PSIGAN & 26.96 & 29.70 & 23.47 & 27.74 & 26.97 & 36.15 & 66.97 & 47.25 & 52.99 & 50.84 \\
    GenericSSL & \bf 5.55 & \bf 6.05 & \underline{4.52} & 23.29 & \underline{9.85} & 37.36 & 47.11 & 39.46 & 52.22 & 44.04 \\
    FPL+ & 12.80 & 17.50 & 9.20 & 18.30 & 14.45 & 28.77 & 65.42 & 38.75 & 42.34 & 43.82 \\
    Diff-style & 17.12 & 88.67 & 22.13 & 50.13 & 44.51 & 53.27 & 43.19 & 55.29 & 43.96 & 48.93  \\ 
    \midrule[0.5pt]
    \textbf{Ours-AD} & \underline{6.78} & 10.50 & \bf 4.38 & \bf 6.80 & \bf 7.11 & \bf 12.59 & 59.19 & \bf 30.18 & 40.39 & 35.59\\
    \textbf{Ours-GS} & 10.20 & \underline{6.07} & 9.27 & 19.31 & 11.21 & 26.96 & \bf 30.77 & 33.57 & 47.97 & \underline{34.82}\\  
    \textbf{Ours-PL} & 7.11 & 11.29 & 5.36 & 17.55 & 10.33 & \underline{22.46} & \underline{40.86} & 43.64 & \bf 27.76 & \bf 33.68 \\
    \bottomrule[0.8pt]
\end{tabular}
}
}
\end{table*}

\begin{table*}[ht]
\centering
\caption{Comparison of different methods on the Abdominal Multi-Organ dataset.}
\label{table_3}
\tiny{
\setlength{\tabcolsep}{0.9mm}{
\begin{tabular}{c|ccccc|ccccc}
    \toprule[0.8pt]
    & \multicolumn{5}{c|}{Abdominal MRI $\rightarrow$ Abdominal CT} & \multicolumn{5}{c}{Abdominal CT $\rightarrow$ Abdominal MRI}\\
    \midrule[0.5pt]
    \multirow{2}{*}{Method} & \multicolumn{5}{c|}{95HD $\downarrow$} & \multicolumn{5}{c}{95HD $\downarrow$} \\
    \cmidrule[0.5pt]{2-11}
    & Liver & R. Kid & L. Kid & Spleen & Average & Liver & R. Kid & L. Kid & Spleen & Average \\
    \midrule[0.5pt]
    No Adaptation & 97.56 & 29.58 & 37.91 & 88.32 & 63.34 & 72.88 & 42.06 & 40.48 & 43.23 & 49.66 \\
    Fully supervised & 14.28 & 11.83 & 6.82 & 9.49 & 10.61 & 14.09 & 10.12 & 17.87 & 25.36 & 16.86 \\
    \midrule[0.5pt]
    CycleGAN & 27.66 & 28.82 & 30.97 & 20.43 & 26.97 & \underline{32.69} & 34.46 & 30.31 & 26.34 & 30.95 \\
    CyCADA & 41.27 & 23.49 & \underline{17.83} & 20.66 & 25.81 & 38.03 & 35.97 & 28.08 & 31.10 & 33.29\\
    ADVENT & 80.76 & 31.86 & 26.98 & 90.31 & 57.48 & 79.51 & 42.06 & 40.48 & 43.65 & 51.42\\
    FDA  & 29.88 & 16.56 & 26.37 & 16.61 & 22.36 & 47.78 & 31.80 & 32.76 & 25.40 & 34.44\\
    SIFA & 30.01 & 37.73 & 18.15 & 32.10 & 29.50 & 41.52 & 52.09 & 25.40 & 45.20 & 41.05 \\
    PSIGAN & 34.66 & 20.41 & 19.12 & 25.91 & 25.03 & 60.94 & 45.34 & 36.84 & 33.58 & 44.17 \\
    GenericSSL & 62.18 & 35.98 & 54.94 & 82.72 & 58.95  & 33.99 & 39.73 & 43.02 & 37.06 & 38.45\\
    FPL+ & 31.40 & 16.90 & 24.50 & 17.60 & 22.60  & 44.35 & 32.75 & 30.70 & 26.40 & 33.55\\
    Diff-style & 62.85 & 21.10 & 17.89 & 19.15 & 30.25 & 70.19 & 41.47 & 34.95 & 37.93 & 46.13\\
    \midrule[0.5pt]
    \textbf{Ours-AD} & \underline{20.74} & 12.45 & 22.16 & \underline{8.82} & \underline{16.04} & 35.94 & 27.95 & 26.93 & \underline{19.84} & 27.66 \\
    \textbf{Ours-GS} & 26.23 & \bf 10.17 & 19.85 & 13.20 & 17.36 & \bf 29.69 & \underline{23.65} & \bf 19.94 & 21.22 & \underline{23.63} \\
    \textbf{Ours-PL} & \bf 17.31 & \underline{10.57} & \bf 14.35 & \bf 7.26 & \bf 12.37 & 35.25 & \bf 19.44 & \underline{22.17} & \bf 14.16 & \bf 22.76\\
    \bottomrule[0.8pt]
\end{tabular}
}
}
\end{table*}

\section{Extra Results on Brats from T2 to Flair}
We provide additional results on the BraTS dataset for the domain adaptation task from T2 to FLAIR. As shown in~\cref{table_t2w_t2f}, our UPDiff-UDA method outperforms all baseline approaches in both Dice score and 95HD, demonstrating its superior segmentation performance.

\begin{table*}[t]
\centering
\caption{Comparison of different methods on the BraTS dataset.}
\scriptsize
\setlength{\tabcolsep}{1mm}{
\begin{tabular}{c|cccc|cccc}
    \toprule[0.8pt]
    Target domain & \multicolumn{8}{c}{Flair}  \\
    \midrule[0.5pt]
    \multirow{2}{*}{Method} & \multicolumn{4}{c|}{Dice (\%) $\uparrow$} & \multicolumn{4}{c}{95HD $\downarrow$} \\
    \cmidrule[0.5pt]{2-9}
    & WT & TC & ET & Average & WT & TC & ET & Average \\
    \midrule[0.5pt]
    No Adaptation & 76.29 & 60.04 & 34.40 & 56.91 & 16.69 & 25.25 & 35.40 & 25.78 \\
    Fully supervised & 87.33 & 68.18 & 47.81 & 67.78 & 6.12 & 17.78 & 24.19 & 16.03 \\
    \midrule[0.5pt]

    CycleGAN & 74.77 & 59.11 & 32.14 & 55.34 & 22.85 & 29.64 & 33.66 & 28.71  \\
    CyCADA & \bf 80.75 & 61.30 & 29.78 & 57.28 & \bf 12.02 & 23.75 & 37.57 & 24.44  \\
    ADVENT & \underline{75.95} & 60.16 & 37.52 & 57.88 & 14.38 & 22.11 & 26.44 & \underline{20.98} \\
    FDA & 75.46 & \underline{62.85} & 38.44 & \underline{58.92} & 20.67 & 30.86 & 41.47 & 31.00 \\
    SIFA & 69.18 & 58.11 & 39.51 & 55.60 & 19.66 & 38.79 & 49.24 & 35.90\\
    PSIGAN & 74.26 & 57.82 & 36.60 & 56.23 & 16.54 & 26.81 & 34.13 & 25.82\\
    GenericSSL & 41.73 & 52.65 & 28.72 & 41.03 & 56.88 & 34.47 & 40.26 & 43.87  \\
    FPL+  & 74.66 & 54.31 & 36.73 & 55.23 & 20.40 & 30.19 & 41.04 & 30.54\\
    \midrule[0.5pt]
    \textbf{Ours-AD} & 73.74 & 62.30 & \underline{39.80} & 58.61 & \underline{12.44} & \bf 18.58 & \bf 25.42 & \bf 18.81 \\
    \textbf{Ours-GS} & 73.86 & 59.66 & 36.08 & 56.53 & 24.43 & 27.45 & 32.61 & 28.17 \\
    \textbf{Ours-PL} & 71.70 & \bf 65.42 & \bf 41.00 & \bf 59.37 & 22.59 & \underline{19.98} & \underline{25.89} & 22.82\\
    \bottomrule[0.8pt]
\end{tabular}
}
\label{table_t2w_t2f}
\end{table*}

\section{Extra Qualitative Results}
\begin{figure}[t]
    \centering 

   \includegraphics[width=1.01\linewidth]{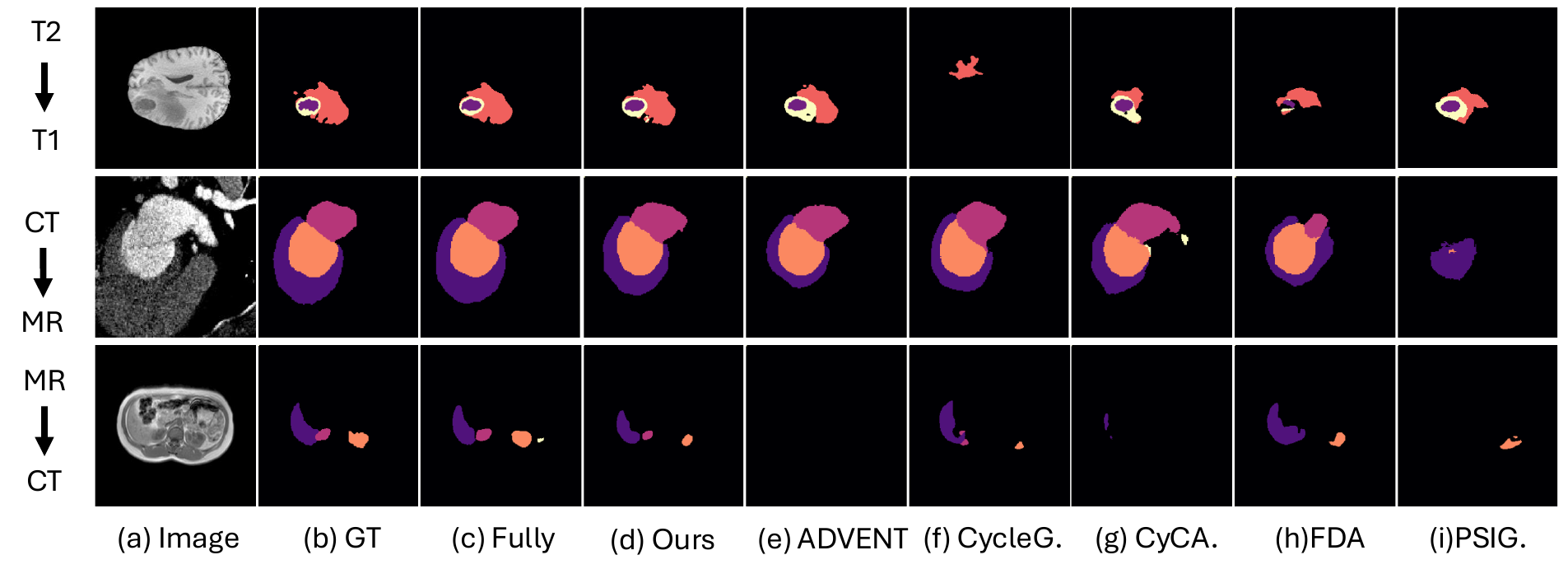}


\caption{Segmentation results of ours (Ours-PL) and other baselines. 
}
\label{qua_res_seg_supplement}
\end{figure}

Here we supply some extra qualitative results in~\cref{qua_res_seg_supplement}. Our method achieves better performance than other baselines.

\section{Ablation study}
From~\cref{alg:udsm}, we know that the confidence threshold $\delta$ is important for the conditional generation of high-quality target images. Here we study the effect of $\delta$ in augmenting domain adaptation training. Our empirical results (\cref{hyper}) show that our method is robust in different values of $\delta$. As described in~\cref {pl_details_supple}, hyperparameter $\lambda$ is the loss weight of pseudo-labeling supervision. The hyperparameters $\tau_l$ and $\tau_u$ are the lower and upper bounds of the confidence threshold for filtering out low-quality pseudo-labels during training. Due to the high-quality generations in the target domain, the performance of the pseudo-labeling framework is robust to the perturbation of $\lambda$, $\tau_l$, and $\tau_u$. This reflects the ability of our UPDiff-UDA in conditionally generating high-quality target domain images with pseudo conditions.  

We present experiments to examine the impact of backbone architecture on final performance. As shown in \cref{architecture}, our method remains robust across different U-Net backbones.
\begin{table}[h]
\centering
\caption{Ablation study results of hyperparameters on the BraTS dataset. Overstriking hyperparameter values is our setting.}
\label{hyper}
\scriptsize{
\setlength{\tabcolsep}{1mm}{
\begin{tabular}{c|cccc|cccc}
    \toprule[0.8pt]
    Target domain & \multicolumn{8}{c}{T1C} \\
    \midrule[0.5pt]
    \multirow{2}{*}{Hyper-parameters} & \multicolumn{4}{c|}{Dice (\%) $\uparrow$} & \multicolumn{4}{c}{Hausdorff Distance (mm) $\downarrow$} \\
    \cmidrule[0.5pt]{2-9}
    & WT & TC & ET & Average & WT & TC & ET & Average  \\
    \midrule[0.5pt]
    $\delta$-0.7 & \bf 62.52 & 66.30 & 42.06 & 56.96 & 25.25 & 36.54 & 41.54 & 34.45 \\
    $\delta$-\textbf{0.8} & 62.47 & \bf 67.89 & \bf 44.12 & \bf 58.16 & \bf 20.89  & \bf 23.07  & \bf 27.84 & \bf 23.93\\
    $\delta$-0.9 & 60.15 & 63.41 & 38.75 & 54.10 & 30.06 & 38.54 & 44.71 & 37.77 \\
    \midrule[0.5pt]
    $\lambda$-0.6 & 62.69 & 66.57 & 42.86 & 57.37 & 25.19 & 30.95 & 35.87 & 30.67 \\
    $\lambda$-\textbf{0.7} &62.47 & \bf 67.89 & \bf 44.12 & \bf 58.16 & \bf 20.89  & \bf 23.07  &  \bf 27.84 & \bf 23.93\\
    $\lambda$-0.8 & \bf 63.96 & 67.45 & 39.69 & 57.03 & 24.57 & 29.14 & 34.28 & 29.33\\
    \midrule[0.5pt]
    $\tau_l$-0.4 & \bf 62.94 &  \bf 68.11 & 39.01 & 56.69 & 25.24 & 26.31 & 31.61 & 27.72 \\
    $\tau_l$-\textbf{0.5} &62.47 & 67.89 & \bf 44.12 & \bf 58.16 & \bf 20.89  & \bf 23.07  & \bf 27.84 & \bf 23.93\\
    $\tau_l$-0.6 & 62.13 & 67.16 & 40.68 & 56.65 & 21.05 & 24.12 & 30.36 & 25.18 \\
    \midrule[0.5pt]
    $\tau_u$-0.4 & 61.24 & 65.01 & 39.03 & 55.09 & 24.92 & 32.31 & 37.20 & 31.48 \\
    $\tau_u$-\textbf{0.7} &62.47 & 67.89 & \bf 44.12 & \bf 58.16 & 20.89  & \bf 23.07  & \bf 27.84 & \bf 23.93\\
    $\tau_u$-0.8 & \bf 62.90 & \bf 67.94 & 40.39 & 57.07 & \bf 18.44 & 25.97 & 32.16 & 25.52\\
    \bottomrule[0.8pt]
\end{tabular}
}
}
\end{table}





\label{sec:ablation_appendix}

\begin{table}[ht]
\centering
\caption{Ablation study on the U-Net architecture.}
\label{architecture}
\scriptsize{
\setlength{\tabcolsep}{1mm}{
\begin{tabular}{c|cccc|cccc}
    \toprule[0.8pt]
    Target domain & \multicolumn{8}{c}{T1C} \\
    \midrule[0.5pt]
    \multirow{2}{*}{Hyper-parameters} & \multicolumn{4}{c|}{Dice (\%) $\uparrow$} & \multicolumn{4}{c}{Hausdorff Distance (mm) $\downarrow$} \\
    \cmidrule[0.5pt]{2-9}
    & WT & TC & ET & Average & WT & TC & ET & Average  \\
    \midrule[0.5pt]
    ours-ResNet18 & 60.12 & 63.32 & 36.93 & 53.46 & 24.48 & 40.51 & 44.17 & 36.38 \\
    ours-ResNet34 &  62.47 & \bf 67.89 & \bf 44.12 & \bf 58.16 & 20.89 & \bf 23.07 & \bf 27.84 & \bf 23.93 \\
    ours-ResNet50 & \bf 62.56 & 66.11 & 42.83 & 57.17 & \bf 20.32 & 25.54 & 31.78 & 25.88  \\
    \bottomrule[0.8pt]
\end{tabular}
}
}
\end{table}

\clearpage
\newpage

\end{document}